\RequirePackage{fix-cm}

\documentclass[smallextended,a4paper]{svjour3}       % onecolumn (second format)
\def\makeheadbox{\relax}
\smartqed  % flush right qed marks, e.g. at end of proof
 \usepackage{mathptmx}      % use Times fonts if available on your TeX system
\usepackage{soul}
\usepackage{array}
\usepackage{arydshln}

\usepackage{lineno,hyperref}
\usepackage{graphicx}
\usepackage[cmex10]{amsmath}
\usepackage{amssymb}
\usepackage{url}
\usepackage{bbm}
\usepackage{textcomp}
\usepackage{inputenc}
\usepackage{multirow}
\usepackage{tabularx}
\usepackage{subfigure}
\usepackage{booktabs}
\usepackage[table]{xcolor}
\usepackage{authblk}
\usepackage[colorinlistoftodos]{todonotes}
\usepackage{adjustbox}
\graphicspath{{fig/}}
\newcolumntype{R}[2]{%
    >{\adjustbox{angle=#1,raise={3pt}{27pt}{4pt}}\bgroup}%
    l%
    <{\egroup}%
} %lap=\width-(#2),
\newcommand*\rot{\multicolumn{1}{R{45}{0em}}}% no optional argument here, please!
\definecolor{lightgray}{gray}{0.9}

\newcommand{\R}[1]{\mathbb{R}^{#1}}
\newcommand{\lm}{\lambda}
\newcommand{\Glx}{G_{\lm}^X}

\newcommand{\1}[1]{\text{\textlbrackdbl}{#1}\text{\textrbrackdbl}}
\newcommand{\T}{^\top}

\newcommand{\h}[1]{{#1}}

\begin{document}

\title{Aggregating Binary Local Descriptors\\ for Image Retrieval
%\thanks{Grants or other notes
%about the article that should go on the front page should be
%placed here. General acknowledgments should be placed at the end of the article.}
}
%\subtitle{Do you have a subtitle?\\ If so, write it here}

%\titlerunning{Short form of title}        % if too long for running head

\author{Giuseppe Amato $\cdot$ Fabrizio Falchi $\cdot$ Lucia Vadicamo}

%\authorrunning{Giuseppe Amato \and Fabrizio Falchi \and Lucia Vadicamo} % if too long for running head

\institute{G. Amato, F. Falchi, L. Vadicamo \at Institute of Information Science and Technologies (ISTI) - CNR \\
Via Moruzzi 1, 56124 Pisa (Italy)\\
\email{\{firstname\}.\{lastname\}@isti.cnr.it}}
%\institute{F. Author \at
%              first address \\
%              Tel.: +123-45-678910\\
%              Fax: +123-45-678910\\
%              \email{fauthor@example.com}           %  \\
%%             \emph{Present address:} of F. Author  %  if needed
%           \and
%           S. Author \at
%              second address
%}

\date{}
% The correct dates will be entered by the editor

\maketitle

\maketitle
\begin{abstract}
\h{Content-Based Image Retrieval based on local features is computationally expensive because of the complexity of both extraction and matching of local feature. On one hand, the cost for extracting, representing, and comparing local visual descriptors has been dramatically reduced by recently proposed binary local features. On the other hand, aggregation techniques provide a meaningful summarization of all the extracted feature of an image into a single descriptor, allowing us to speed up and scale up the image search. Only a few works have recently mixed together these two research directions, defining aggregation methods for binary local features, in order to leverage on the advantage of both approaches.}
In this paper, we report an extensive comparison among state-of-the-art aggregation methods applied to binary features.
Then, we mathematically formalize the application of Fisher Kernels to Bernoulli Mixture Models.
Finally, we investigate the combination of the aggregated binary features with the emerging Convolutional Neural Network (CNN) features.
Our results show that aggregation methods on binary features are effective and represent a worthwhile alternative to the direct matching. 
Moreover, the combination of the CNN with the Fisher Vector (FV) built upon binary features allowed us to obtain a relative improvement over the CNN results that is in line with that recently obtained using the combination of the CNN with the FV built upon SIFTs.  
The advantage \h{of using the FV built upon binary features} is that the extraction process of binary features is about two order of magnitude faster than SIFTs.
%In this paper, we report an extensive comparison among state-of-the-art aggregation methods applied to binary features.
%Then, we mathematically formalize the application of Fisher Kernels to Bernoulli Mixture Models.
%Finally, we investigate the combination of the aggregated binary features with the emerging Convolutional Neural Network (CNN) features.
%Our results show that aggregation methods on binary features are effective and represent a worthwhile alternative to the direct matching. 
%Moreover, the combination of the CNN with the Fisher Vector (FV) built upon binary features allowed us to obtain a relative improvement over the CNN results that is in line with that recently obtained using the combination of the CNN with the FV built upon SIFTs.  
%The advantage is that the extraction process of binary features is about two order of magnitude faster than SIFTs.

\end{abstract}

 %, Information Search and Retrieval
  %Bernoulli mixture Model
 
 \section{Introduction}\label{sec:introduction}
Content-Based Image Retrieval (CBIR) is a relevant topic
studied by many scientists in the last decades. CBIR refers to the
possibility of organizing archives containing digital pictures, so
that they can be searched and retrieved by using their visual
content \cite{datta05}. A specialization of the basic CBIR
techniques include the techniques of object recognition
\cite{Ullman96}, where visual content of images is analyzed so
that objects contained in digital pictures are recognized, and/or
images containing specific objects are retrieved. Techniques of
CBIR and object recognition are becoming increasingly popular in
many web search engines, where images can be searched by using
their visual content \cite{Google-Images,Bing-Images}, and on
smartphones apps, where information can be obtained by pointing the
smartphone camera toward a monument, a painting, a logo
\cite{Google-Goggles}.

During the last few years, \textit{local descriptors}, as for instance SIFT
\cite{lowe04}, SURF \cite{bay06}, BRISK \cite{leutenegger11}, ORB
\cite{rublee11}, to cite some, have been widely used to support
effective CBIR and object recognition tasks. A local descriptor is
generally a histogram representing statistics of the pixels in
the neighborhood of an interest point (automatically) chosen in an
image. %An image typically contains thousands of interest points, and correspondingly, thousands of local descriptors. 
Among the
promising properties offered by local descriptors, we mention the
possibility to help mitigating the so called \emph{semantic gap}
\cite{Smeulders:2000:CIR:357871.357873}, that is the gap between
the visual representation of images and the semantic content of
images. In most cases visual similarity does not imply semantic
similarity.

Executing image retrieval and object recognition tasks, relying on
local features, is generally resource demanding. Each digital
image, both queries and images in the digital archives, are
typically described by thousands of local descriptors.  In order
to decide that two images match, since they contain the same or
similar objects, local descriptors in the two images need to be
compared, in order to identify matching patterns. This
poses some problems when local descriptors are used on devices
with low resources, as for instance smartphones, or when response
time must be very fast even in presence of huge digital archives.
On one hand, the cost for extracting local descriptors, storing
all descriptors of all images, and performing feature matching
between two images must be reduced to allow their interactive use
on devices with limited resources. On the other hand, compact
representation of local descriptors and ad hoc index structures
for similarity matching \cite{ZADB06Similarity} are needed to
allow image retrieval to scale up with very large digital picture
archives. These issues have been addressed by following two
different directions.

To reduce the cost of extracting, representing, and matching local
visual descriptors, researchers have investigated the use\textit{ binary
local descriptors}, as for instance BRISK %\cite {leutenegger11} 
and
ORB. %\cite{rublee11}.
Binary features are built from a set of pairwise intensity comparisons.
Thus, each bit of the descriptors is the result of exactly one comparison.
Binary descriptors are much faster to be extracted, are
obviously more compact than non-binary ones, and can also be
matched faster by using the Hamming distance
\cite{Hamming:1950:EDE} rather than the Euclidean distance.
 For example, in \cite{rublee11} it has been showed that ORB is an 
 order of magnitude faster than SURF, and over two orders faster than SIFT.
However, note that even if binary local descriptors are compact,
each image is still associated with thousand local descriptors,
making it difficult to scale up to very large digital archives.

%Reduction of the cost of image matching on a very large scale has
%been obtained by using methods for quantizing and/or
%aggregating local features. 
\h{The use of the information provided by each local feature is crucial for tasks such as image stitching and 3D reconstruction. For other tasks such as image classification and retrieval, high effectiveness have been achieved using the \textit{quantization} and/or \textit{aggregation techniques} which provide meaningful summarization of all the extracted features of an image} \cite{jegou10:VLAD}\h{.
One profitable outcome of using quantization/aggregation techniques is that they allow us to represent an image by a single descriptor rather than thousands descriptors. This reduces the cost of image comparison and leads to scale up the search to large database.}
On one hand, quantization methods, as
for instance the Bag-of-Words approach (BoW) \cite{sivic03}, %bag of feature approach (BoF) \cite{sivic03},
define a finite vocabulary of ``visual words", that is a finite
set of local descriptors to be used as representative. Every
possible local descriptors is thus represented by its closest
visual word, that is the closest element of the vocabulary. In
this way images are described by a set (a bag) of identifiers of
representatives, rather than a set of histograms. On the other
hand, aggregation methods, as for instance Fisher Vectors (FV)
\cite{perronnin07} or Vectors of Locally Aggregated Descriptors
(VLAD) \cite{jegou10:VLAD}, analyze the local descriptors
contained in an image to create statistical summaries that still
preserve the effectiveness power of local descriptors and allow
treating them as global descriptors. In both cases index
structures for approximate or similarity matching
\cite{ZADB06Similarity} can be used to guarantee scalability on
very large datasets. 

\h{Since quantization and aggregation methods are defined and used almost
exclusively in conjunction with non-binary features, the cost of extracting
local descriptors and to quantize/aggregate them on the fly, is
still high.} 
%However, given that these methods are
%basically defined on non-binary features, the cost of extracting
%local descriptors and to quantize/aggregate them on the fly, is
%still high, for some applications.
\h{Recently, some approaches that attempt to integrate the binary
local descriptors with the quantization and aggregation methods
have been proposed in literature}  \cite{galvez11,grana13,lee15,van14,uchida13,zhang13}. \h{In these proposals, the aggregation is directly applied on top of binary local
descriptors.
The objective is to improve efficiency and reduce computing resources needed for image matching  by leveraging on the advantages of
both aggregation techniques (effective compact image representation) and binary local features (fast feature extraction), by reducing, or eliminating the disadvantages.}
%\h{The objective is to improve efficiency by leveraging on the advantages of
%both aggregation techniques (effective compact image representation) and binary local features (fast feature extraction).}

%The objective is to leverage on the advantages of both aggregation techniques and binary local features, by reducing, or eliminating the disadvantages.

%However, given that these methods are
%basically defined on non-binary features, the cost of extracting
%local descriptors and to quantize/aggregate them on the fly, is
%still high, for some applications.
%Recently, some approaches that attempt to integrate the binary
%local descriptors with the quantization and aggregation methods
%have been proposed in literature. In these proposals, aggregation and
%quantization methods have been directly applied on top of binary local
%descriptors. The objective is to leverage on the advantages of
%both approaches, by reducing, or eliminating the disadvantages.

The contribution of this paper is
providing an extensive comparisons and analysis of the aggregation and quantization methods applied to binary local descriptors also providing a novel formulation of Fisher Vectors built using the Bernoulli Mixture model (BMM)\h{, referred to as BMM-FV.}
Moreover, we investigate the combination of \h{BMM-FVs and other encodings of} 
%FVs built upon 
binary features with the Convolutional Neural Network \cite{razavian2014cnn} features as other case of use of binary feature aggregations.
% We do not directly compare with recent methods based on
% Convolutional Neural Network techniques \cite{razavian2014cnn} since our aim is not obtaining overall best effectiveness.
%In fact, 
We  focus on cases where, for efficiency issues \cite{rublee11,heinly12}, the binary features are extracted and used to represent images. Thus, we compare aggregations of binary features in order to find the most suitable techniques to avoid the direct matching.
We expect this topic to be relevant for application that uses binary features on devices with low CPU and memory resources, as for instance mobile and
wearable devices. In these cases the combination of aggregation
methods with binary local features is very useful and led to scale up image search on large scale, where direct matching is not feasible.
% However, on the basis of some recent results \cite{chandrasekhar2015, amato16:JOCCH} where the information provided by the FV built upon SIFTs has been used to further improve the retrieval performance of the CNN feature, in this paper we also investigates the combination of FVs built upon binary features with the CNN features as other case of use of binary feature aggregations.

This paper extends our early work on aggregations of binary features  \cite{amato16} by 
a) providing a  formulation of the Fisher Vector %(BMM-FV) 
built using the Bernoulli Mixture Model (BMM) which preserve the structure of the traditional FV built using a Gaussian Mixture Model (existing implementations of the FV can be easily adapted to work also with BMMs);
b) comparison of the BMM-FV against the other state-of-the-art aggregation approaches on two standard benchmarks (INRIA Holidays\footnote{Respect to the experimental setting used in our previous work\h{ }\cite{amato16}, we improved the computation of the local features before the aggregation phase which allowed us to obtain better performances for BoW and VLAD on the INRIA Holidays dataset than that  reported in \cite{amato16}.} \cite{jegou08} and {Oxford5k }\cite{philbin07});
c) evaluation of the BMM-FV on the top of several binary local features (ORB \cite{rublee11}, LATCH \cite{levi15_LATCH}, AKAZE \cite{akaze}) whose performances have not been yet reported on benchmark for image retrieval;
d) evaluation of the combination of the BMM-FV with the emerging Convolutional Neural Network (CNN) features, \h{including experiments on a large scale}. % to improve the latter retrieval performance.
 The results of our experiments show that the use of aggregation
and quantization methods with binary local descriptors is
generally effective even if, as expected, retrieval performance is worse than that obtained applying the same aggregation and quantization
methods directly to non-binary features.
%Therefore, the loss of information due to the double approximation (binarization+aggregation/quantization) is limited.
The BMM-FV approach provided us with performance results that are
better than all the other aggregation methods on binary descriptors. In addition, our results show that some aggregation methods led to obtain very compact image representation with a retrieval performance comparable to the direct matching, which actually is the most used approch to evaluate the similarity of images described by binary local features.
Moreover, we show that the combinations of BMM-FV and CNN improve the latter retrieval performances and achieves effectiveness comparable with that obtained combining CNN and FV built upon SIFTs, previous proposed in \cite{chandrasekhar2015}. The advantage \h{of combining BMM-FV and CNN instead of combining traditional FV and CNN} is that BMM-FV relies on binary features whose extraction is noticeably faster than SIFT extraction. 

The paper is organized as follows. Section \ref{sec:rel} offers an
overview of other articles in literature, related to local features,
binary local features, and aggregation methods. Section
\ref{sec:imagerep} discusses how existing aggregation methods
can be used with binary local features. It also contains our
approach for applying Fisher Vectors on binary local features and how combining it with the CNN features. 
Section \ref{sec:exp} discusses the evaluation experiments and the
obtained results. Section \ref{sec:conclusion} concludes.

 \section{Related Work} \label{sec:rel}
 %%PRESENT PERFECT: to refer to ongoing situations, i.e. when authors are still investigating a particular field.
 %%PAST SIMPLE must be used when you are talking about completely finished actions:
 %%          -The year of publication is stated within the main sentence (i.e. not just in brackets)
 %%          -You mention specific pieces of research (e.g. initial approaches and methods that have subsequently probably been abandoned)
 %%          -You state the exact date when something was written, proved etc.
 %%SIMPLE PRESENT:descriptions of established scientific fact+
 %%              describing how a system, method, procedure etc. functions
 %%              to discuss previously published laws, theorems, definitions,proofs, lemmas etc.
 %%Esempi:
 %%Lindley [10] INVESTIGATED the use of the genitive in French and English and his results AGREE with other authors? findings in this area [12, 13, 18]. He PROVED that ?
 %%In [5] Evans STUDIES/STUDIED the differences
 %%The theorem STATES that the...
 
  The research for effective representation of visual feature for
  images has received much attention over the last two decades.
  %LF e binary
 The use of \textit{local features}, such as SIFT \cite{lowe04} and SURF
  \cite{bay06}, is at the core of many computer vision applications,
  since it allows systems to efficiently match local structures between
  images. 
 % The local features are numerical representation of the visual appearance of local regions of an image. 
  %The extraction process can be divided in two phases: first, a set of interest points, referred to as \textit{keypoints}, are automatically detected; then one or more descriptors are associated to each keypoint.
   %
      To date, the most used and cited local feature is the Scale Invariant Feature Transformation (SIFT) \cite{lowe04}. The success of SIFT is due to its distinctiveness that enable to effectively find correct matches between images. However, the SIFTs extraction is costly due to the local image gradient computations. % which makes it unsuitable for time-constrained applications, e.g. real-time recognition. %involves
    In \cite{bay06} integral images were used to speed up the computation and the SURF feature was proposed as an efficient approximation of the SIFT. 
    To further reduce the cost of extracting, representing, and matching
    local visual descriptors, researchers have investigated the \textit{binary local descriptors}. %, as of instance ORB \cite{rublee11} and BRISK\cite{leutenegger11}.
    These features have a compact binary representation that is not the result of a quantization, but rather is computed directly from pixel-intensity comparisons.
     One of the early studies in this direction was the Binary Robust Independent Elementary Features (BRIEF) \cite{calonder10}. %Each BRIEF descriptor is computed by comparing a sample set of pixel pairs of a smoothed image patch. For each pair, if the first pixel intensity is larger than the second, the corresponding entry in the BRIEF descriptor is 1 and 0 otherwise. 
	Rublee et al.\h{ }\cite{rublee11} proposed a %very fast 
    binary feature, called ORB (Oriented FAST and Rotated BRIEF), whose extraction process is an order of magnitude faster than SURF, and two orders faster than SIFT according to the experimental results reported in \cite{rublee11,Miksik2012,heinly12}.
    %In fact, in the experiments reported in \cite{rublee11}, the time required to compute about $1,000$ features from an image was about $15$ ms for ORB and more than $1.5$ second for SIFT. Similar results were reported also in \cite{Miksik2012,heinly12}.
     Recently, several other binary local features have been proposed, such as BRISK \cite{leutenegger11}, AKAZE \cite{akaze}, and LATCH \cite{levi15_LATCH}.
     
     Local features have been widely used in literature and applications, however since each image is represented by thousands
          of local features there is a significant amount of memory consumption and time  required to compare local features  within large databases.
  \h{
\textit{Aggregation techniques} have been introduced to summarize the information contained in all the local features extracted from an image into a single descriptor. The advantage is twofold: 1) reduction of the cost of image comparison (each image is represented by a single descriptor rather than thousands descriptors); 2) aggregated descriptors have been proved to be particularly effective for image retrieval and classification task.}
 
%    The use of the information provided by each local feature is crucial for some tasks such as image stitching and 3D reconstruction. For other tasks such as image classification and retrieval high effectiveness have been achieved using the \textit{aggregation techniques} which provide meaningful summarization of all the extracted feature of an image. % \cite{jegou10:VLAD}.
%     One profitable outcome of using aggregation techniques is that each image is represented by a single descriptor rather than thousands descriptors. This reduces the cost of image comparison and leads to scale up the search to large database.
   
    %Please note that even if binary local descriptors are faster and more compact respect to non-binary one, each image is still associated with thousand local descriptors, making it difficult to scale up the search to very large digital archives.
    %%The execution of image retrieval and object recognition tasks, relying on local features (binary or not), is generally resource demanding. In fact, each digital image, both queries and images in the digital archives, are typically described by thousands of local descriptors. 
    %In fact, in order to decide that two images match (i.e., contain the same or similar objects), local descriptors in the two images need to be pairwise compared to identify matching patterns.
    %To reduce the cost of image matching on a very large scale methods for aggregating local features have been proposed, as for example Bag-of-Words \cite{sivic03} and Fisher Vector \cite{perronnin07}. 

 %%AGGREGATIOM
  %% BoW
  By far, the most popular aggregation method has been the {Bag-of-Word} (BoW) \cite{sivic03}. BoW  was initially proposed for matching object in video and has been studied in many other papers, such as  \cite{csurka04,philbin07,jegou10:improvingBoW,jegou10:VLAD}, for classification and CBIR tasks.
  BoW uses a visual vocabulary to quantize the local descriptors extracted from images; each image is then represented by a histogram of occurrences of visual words.
  The BoW approach used in computer vision is very similar to the BoW used in natural language processing and information retrieval \cite{salton86}, thus many text indexing techniques, such as inverted files \cite{witten99}, have been applied for image search.
%   %However, as highlighted in \cite{zhang09},
%%  an image query contains many more terms than a text query, e.g. 1500 visual terms rather than 3 text terms,
%%  thus the inverted files are not completely adequate to index images 
%%  (thousand posting list should be accessed).
%  From the very beginning %\cite{sivic03} 
%  words reductions techniques have been used and images have been ranked using the standard \emph{term frequency-inverse document frequency} ({tf-idf})  weighting. %\cite{salton86}
%%  In the TOP-SURF \cite{thomee10} representation  just the highest scoring visual words have been used and in \cite{amato13:onReducing} several approaches for the reduction of visual words have been investigated, in order to improve the efficiency %of the BoW
%%  with a slight loss in effectiveness. 
  Search results obtained using BoW in CBIR have been improved by exploiting additional geometrical information \cite{philbin07,perdoch09,tolias11:SpeededUp,zhao13}, applying re-ranking approaches \cite{philbin07,jegou08,chum07,tolias13:queryExp} or using better encoding techniques, such as the {Hamming Embedding} \cite{jegou08}, {soft/multiple-assignment} \cite{philbin08,vanGemert08,jegou10:improvingBoW}, {sparse coding} \cite{yang09,boureau10}, {locality-constrained linear coding} \cite{wang10} and {spatial pyramids} \cite{lazebnik06}.
 % As others have highlighted \cite{arandjelovic12:rootsift, jegou10:improvingBoW,philbin08}, the quantization process reduces the discriminative power of the local features: the information about the original descriptors is lost and corresponding descriptors may be assigned to different visual words. To overcome the quantization loss, Jegou et al. \cite{jegou08,jegou10:improvingBoW} have exploited a more accurate representation of the original descriptors in the so called {Hamming Embedding} (HE) method. %\cite{jegou10:improvingBoW}.
%  Furthermore, better encoding techniques have been used, such as {soft-assignment} \cite{philbin08,vanGemert08,vanGemert10}, {multiple assignment} \cite{jegou10:improvingBoW,jegou10:AccurateImage}, {sparse coding} \cite{yang09,boureau10}, {locality-constrained linear coding} \cite{wang10} or the use of {spatial pyramids} \cite{lazebnik06}.

 %%FV
  Recently, alternative encodings schemes, like the Fisher Vectors (FVs) \cite{perronnin07} and the Vector of Locally Aggregated Descriptors (VLAD) \cite{jegou10:VLAD}, have attracted much attention because of their effectiveness in both image classification and large-scale image search. The FV uses the Fisher Kernel framework \cite{jaakkola98} to transform an incoming set of descriptors into a fixed-size vector representation. %, that is compatible with the cosine similarity. 
  The basic idea is to characterize how a sample of descriptors deviates from an average distribution that is modeled by a parametric generative model. 
  The Gaussian Mixture Model (GMM) \cite{mclachlan2000} %, estimated on a training set, 
  is typically used as generative model and might be understood as a ``probabilistic visual vocabulary". 
  While BoW counts the occurrences of visual words and so takes in account just 0-order statistics, the FV offers a more complete representation by encoding higher order statistics (first, and optionally second order)
  related to the distribution of the descriptors.
  %While BoW counts the number of occurrences of visual words and so it takes in account just 0-order statistic related to the distribution of the descriptors, the Fisher Vector offers a more complete representation of the sample set by encoding higher order statistics (first, and optionally second order).
  %The better use of the information provided by the
  The FV results also in a more efficient representation, since fewer visual words are required in order to achieve a given performance.
  However, the vector representation obtained using BoW is typically quite sparse while that obtained using the Fisher Kernel is almost dense. This leads to some storage and input/output issues that have been addressed by using techniques of dimensionality reduction, such as \h{the} Principal Component Analysis (PCA) \cite{bishop06}, compression with product quantization \cite{gray98,jegou11:PQ} and binary codes \cite{perronnin10}.
  In \cite{chandrasekhar2015} a fusion of FV and CNN  features\h{ }\cite{razavian2014cnn,DeCaf} was proposed %.  
  %Recently, 
  and other works \cite{perronnin2015,Simonyan2013,Sydorov2014} have started exploring the combination of FVs and CNNs by defining hybrid architectures. %For example, in \cite{perronnin2015} Perronin et al. proposed a 
%%  The Fisher Kernel was initially applied for classifying DNA splice site sequences and to detect homologies between protein sequences \cite{jaakkola98}.  Then it has been applied in various contexts as large-scale web audio classification \cite{moreno2000}, speaker classification and identification \cite{fine2001,wan2003}, document classification \cite{vinokourov2001}, object recognition \cite{holub2005} and image classification \cite{perronnin07}.
%%      Perronin el a. \cite{perronnin07} applied the Fisher Kernel in conjunction with a visual vocabulary represented by the means of a Gaussian Mixture Model and showed that the Fisher Kernel can been used
%%      %The Fisher kernel has been applied in the context of
%%      %image classification \cite{perronnin07} and %large scale image search \cite{perronnin10}. It % In image search framework
%%      %it has been used
%%      as efficient tool to aggregate image local descriptors into a fixed-size vector representation. %\cite{jegou12}.
%%     
  
  %%VLAD
  The VLAD method, similarly to BoW, starts with the quantization of
  the local descriptors of an image by using a visual vocabulary
  learned by $k$-means. Differently from BoW, VLAD encodes the
  accumulated difference between the visual words and the associated
  descriptors, rather than just the number of descriptors assigned
  to each visual word. Thus, VLAD exploits more aspects of the
  distribution of the descriptors assigned to a visual word. As
  highlighted in \cite{jegou12}, VLAD might be viewed as a simplified
  non-probabilistic version of the FV. In the original scheme
  \cite{jegou10:VLAD}, as for the FV, VLAD was $L_2$-normalized.
  Subsequently a power normalization step  was introduced for
  both VLAD and FV \cite{jegou12,perronnin10}. Furthermore, PCA
  dimensionality reduction and product quantization were
  applied and several enhancements to the basic VLAD were
  proposed \cite{arandjelovic13:allAbVALD,chen11,delhumeau13,zhao13}.
%%  Arandjelovi\'c et al. \cite{arandjelovic13:allAbVALD} have improved the performance of the VLAD, first by introducing a new normalization scheme, named intra-normalization, second by using a vocabulary adaptation and finally by recording multiple VLAD representations (MultiVLAD) for each image.  %multiple VLAD representations covering subregions of the image.
%%  Different normalization schemes have been investigated also in \cite{chen11} and \cite{delhumeau13}, while Zhao et al. \cite{zhao13} have proposed a pooling strategies termed Covariant-VLAD (CVLAD).
  
%%  Tolias et al. \cite{tolias13} highlight that there is a relationship between the structure of  BoW, HE, VLAD and FV: all these approaches can be reinterpreted using the framework of match kernel \cite{bo09, vedaldi12}.
%%  Thanks to the generic formulation provided by the match kernel, it is possible to decompose al these methods
%%  into embedding and aggregation steps \cite{jegou14:triangulation}. The embedding step maps each local descriptor into a feature space, while the aggregation phase combines a set of feature vectors into a single vector representation.
%%  In \cite{jegou14:triangulation} an innovative embedding method, named Triangular Embedding,  and a ``democratic" aggregation have been proposed in order to reduce the interference between unrelated descriptors. The aggregation stage has also been improved in \cite{tolias14} by jointly encoding geometric information about the dominant orientations associated with the regions of interest of the image. 
%%  

The aggregation methods have been defined and used almost exclusively
 in conjunction with local features that have a real-valued
 representation, such as SIFT and SURF. Few articles have addressed the problem of modifying the state-of-the-art aggregation methods to work with the emerging binary local features.
 In \cite{galvez11,zhang13,grana13,lee15}\h{ }the use of ORB descriptors was integrated into the BoW model by using different clustering algorithms.
 In \cite{galvez11} the visual vocabulary was calculated by binarizing the centroids obtained using the standard $k$-means. In \cite {zhang13,grana13,lee15}
 %, named $k$-majority, that uses a voting scheme to produce binary centroids. This results in modifying
 the $k$-means clustering was modified to fit the binary features by replacing the Euclidean distance with the Hamming distance, and by replacing the mean operation with the median operation.
 %The possibility of integrating binary features into the bag-of-features model has also been explored in \cite{lee15}
 % C. Whiten, R. Laganiere, and G.-A. Bilodeau.Efficient action recognition with MoFREAK. In CRV,pages 319?325, 2013
 In \cite{van14} the VLAD image signature was adapted to work with  binary descriptors: % both $k$-means and $k$-majority
 $k$-means is %algorithm is tested
 used for learning the visual vocabulary and the VLAD vectors are computed in conjunction with an intra-normalization and a final binarization step. %An extentions of the FV to binary features has been carried in \cite{uchida13} were the Bernoulli Mixture Model is used rather then the Gaussian Mixture Model.
  Recently, also the FV scheme has been adapted for the use with binary descriptors: Uchida et al. \cite{uchida13} derived a FV where the Bernoulli Mixture Model \h{ }was used instead of the GMM to model binary descriptors, while Sanchez and Redolfi \cite{sanchez15} generalized the FV formalism to a broader family of distributions, known as the exponential family, that encompasses the Bernoulli distribution as well as the Gaussian one.

 \section{Image Representations}\label{sec:imagerep}
  In order to decide if two images contain the same object or have a similar visual content, one needs an appropriate mathematical description of each image. In this section, we describe some of the most prominent approaches to transform an input image into a numerical descriptor. 
  First we describe the principal aggregation techniques and the application of them to binary local features. Then, the emerging CNN features are presented.
 
 \subsection{Aggregation of local features}\label{sec:aggregation}
 In the following we review how quantization and aggregation
 methods have been adapted to cope with binary features.
 Specifically we present the BoW  \cite{sivic03}, the VLAD
 \cite{jegou10:VLAD} and the FV \cite{perronnin07} approaches.
 \subsubsection{Bag-of-Words}
% The Bag of (Visual) Words (BoW) was initially proposed in \cite{sivic03} for matching objects throughout a video database. %The approach was inspired by the BoW model used in text retrieval.
% Thereafter, it has been widely used for classification and CBIR tasks \cite{csurka04,jegou10:improvingBoW}. 
%
 %In a nutshell, BoW 
 {The Bag of (Visual) Words (BoW)} \cite{sivic03}
 uses a visual vocabulary to group together the local descriptors of an image and represent each image as a set (bag) of
 visual words.   The visual vocabulary is built by clustering the local descriptors of a dataset, e.g. by using $k$-means \cite{kmeans}. The cluster centers, named \emph{centroids}, act as the \textit{visual words} of the vocabulary and they are used to quantize the local descriptors extracted from the images. Specifically, each local descriptor of an image is assigned to its closest centroid and the image is represented by a histogram of occurrences of the visual words.
  The retrieval phase is performed using text retrieval techniques, where visual words are used in place of text word and considering a query image as disjunctive term-query.
 Typically, the cosine similarity measure in conjunction with a term weighting scheme, e.g. {term frequency-inverse document frequency} ({tf-idf}), % \cite{salton86}, %,
 is adopted for evaluating the similarity between any two images.
  % The BoF approach can also be used in conjunction with geometry consistency checks typically performed using RANdom SAmple Consensus (RANSAC) [Fischler 1981] to find
 % homography transformations. In this case, the BoF representation allows fast matching between local features. In fact, any two local features assigned to the same
 % visual words are considered to match.
 % However, as mentioned in [Zhang 2009], �a fundamental difference between an image query (e.g. 1500 visual terms) and a text query (e.g. 3 terms) is largely ignored
 % in existing index design�. Efficiency and memory constraints have been recently addressed by aggregating local descriptors into a fixed-size vector representation
 % that describe the whole image. In particular, Fisher Vector (FV) [S�nchez 2013] and VLAD have shown better performance than BoF. In this work, we will focus on VLAD
 % which has been proved to be a simplified non-probabilistic version of FV [J�gou 2012]. Despite its simplicity, VLAD effectiveness is comparable to that of FV while,
 % in terms of efficiency VLAD is preferable.
 \paragraph{BoW and Binary Local Features}
  In order to extend the BoW scheme to deal with binary features we need a cluster algorithm able to deal with binary strings and Hamming distance. The \textit{$k$-medoids} \cite{kaufman87} are suitable for this scope, but they requires a computational effort to calculate a full distance matrix between the elements of each cluster. In \cite{grana13} it was proposed to use %a variation of the Lloyd algorithm \cite{qualcosa su Lloyd algorithm}, named \emph{$k$-majority}, that uses
  a voting scheme, named \textit{$k$-majority}, %to compute the centroids of a set of binary vectors based on the Hamming distance. The $k$-majority algorithm
  to process a collection of binary vectors and seek for a set of good centroids, that will become the visual words of the BoW model. 
%  Initially the centroids are randomly selected and each element of the  collection is associated with the nearest centroid (according to the Hamming distance).
%  Subsequently, for each cluster, a new centroid is computed by assigning 1 to its $i$-th bit if the majority of the binary vectors in the cluster have a 1 in the  $i$-th bit.
%  The algorithm iterates until no centroids are changed during the previous iteration.
%  After that, %the visual words are computed,
%  the BoW aggregation is performed in the usual manner, by using the Hamming distance rather than the Euclidean. %Also
  An equivalent representation is given also in \cite{zhang13,lee15}, where the BoW model and the $k$-means clustering have been modified to fit the binary features by replacing the Euclidean distance with the Hamming distance, and by replacing the mean operation with the median operation. %The resulting representation is equivalent to the BoW based on $k$-majority.
 \subsubsection{Vector of Locally Aggregated Descriptors}
The Vector of Locally Aggregated Descriptors (VLAD) was initially proposed in \cite{jegou10:VLAD}. As for the BoW, a visual vocabulary $\{\mu_1, \ldots, \mu_K\}$ is first learned using a clustering algorithm (e.g. $k$-means).
Then each local descriptor $x_t$ of a given image is  associated with its nearest visual word $NN(x_t)$ in the vocabulary and for each %codeword
centroid $\mu_i$ the differences $x_t-\mu_i$ of the vectors $x_t$ assigned to $\mu_i$ are accumulated:
%\begin{equation*}
$v_i=\sum_{x_t:NN(x_t)=\mu_i} x_t - \mu_i.$
%\end{equation*}
The VLAD is the concatenation of the residual vectors $v_i$, i.e. $V=[v_1\T \ldots v_K\T]$.
All the residuals have the same size $D$ which is equal to the size of the used local features.
Thus the dimensionality of the whole vector $V$ %describing any image
is fixed too and it is equal to $DK$.

%Power-law and $L_2$ normalization are usually applied and %in order to improve the effectiveness of the VLAD approach .
%Euclidean distance has been proved to be effective for comparing two VLADs.
%VLAD descriptors have high dimensionality; PCA has been used to have a more compact representation \cite{jegou10:VLAD}.
 \paragraph{VLAD and Binary Local Features}
 A naive way to apply the VLAD scheme to binary local descriptors is treating binary vectors as a particular case of real-valued vectors. In this way, the \textit{$k$-means} algorithm can be used to build the visual vocabulary and the difference between the centroids and the descriptors can be accumulated as usual. This approach has also been used in \cite{van14},  where a variation to the VLAD image signature, called BVLAD, has been defined to work with binary features.
 Specifically, the BVLAD is the binarization (by thresholding) of a VLAD obtained using  power-law, intra-normalization, $L_2$ normalization and multiple PCA.  
  Thereafter we have not evaluated the performance of the BVLAD because the binarization of the final image signature is out of the scope of this paper. %In fact, the BVLAD, unlike the other techniques of aggregation of binary features, is obtained using a binarization by thresholding which is an extra step after the aggregation phase.

Similarly to BoW, various binary-cluster algorithms (e.g. \textit{$k$-medoids} and \textit{$k$-majority}) and the  Hamming distance can be used to build the visual vocabulary  and associate each binary descriptor to its nearest visual word. However, as we will see, the use of binary centroid\h{s} may provide less discriminant information during the computation of the residual vectors.% and thus results in less effective retrieval performance. 
% Similarly to BoW, various binary-cluster algorithms (e.g. \textit{$k$-medoids} and \textit{$k$-majority}) and the Hamming distance  can be used to build the visual vocabulary and associate each binary descriptor to its nearest visual word. However, the use of binary centroids may provide less discriminant information during the computation of the residual vectors.
 
 \subsubsection{Fisher Vector}
% The Fisher Kernel is a powerful framework  initially proposed in \cite{jaakkola98}
% for classifying DNA splice site sequences.
% In \cite{perronnin07}, the Fisher Kernel method {was} adopted in the context of image classification as efficient tool to %aggregate
% {encode} image local descriptors into a fixed-size vector representation.
 The Fisher Kernel \cite{jaakkola98} is a powerful framework  adopted  in the context of image classification in \cite{perronnin07} as efficient tool to %aggregate
  {encode} image local descriptors into a fixed-size vector representation.
   The main idea %of this method 
   is to derive a kernel function to measure the similarity between two sets of data, such as the sets of local descriptors extracted from two images. 
   The similarity of two sample sets $X$ and $Y$ is measured by analyzing the difference between the statistical properties of $X$ and $Y$, rather than comparing directly $X$ and $Y$.
   To this scope a probability distribution $p(\cdot|\lm)$ with some parameters $\lm\in\R{m}$  is first estimated on a  training set and it is used as generative model over the 
     the space of all the possible data observations. Then each set $X$ of observations is represented by a vector, named \textit{Fisher Vector}, that indicates the direction in which the parameter $\lambda$ of the probability distribution $p(\cdot|\lm)$ should be modified to best fit the data in $X$.
%   whole sample space $\Omega$ (i.e. the space of all the possible data observations). Then each set $X$ of observations, is represented by a vector, named \textit{Fisher Vector}, that indicates the direction in which the parameter $\lambda$ of the probability distribution $p(\cdot|\lm)$ should be modified to best fit the data in $X$.
   In this way, two samples are considered similar if the directions given by their respective Fisher Vectors are similar.
   Specifically, as proposed in \cite{jaakkola98}, the similarity between two sample sets $X$ and $Y$ is measured using the \textit{Fisher Kernel}, defined as 
%   \begin{equation}
   %K(X,Y)=(\mathcal{G}_{\lm}^X)\Tt  \mathcal{G}_{\lm}^Y,
   $ K(X,Y)=  (\Glx)\T  F_\lm^{-1} G_{\lm}^Y,$
%   \end{equation}
    where $F_\lm$ is the \textit{Fisher Information Matrix} (FIM) and 
    ${G}_{\lm}^X=\nabla_\lm\log p(X|\lm)$ is referred to as the \textit{score function}.
    
    %The vector $\mathcal{G}_{\lm}^X$ is the \emph{Fisher Vector} (FV) of $X$ \cite{perronnin10}. Note that the FV is a fixed size vector whose dimensionality only depends on the dimensionality $m$ of the parameter $\lambda$.  The FV is further divided by $T$ in order to avoid the dependence on the sample size \cite{sanchez13}.
    
   The computation of the Fisher Kernel is costly due the multiplication by the inverse of the FIM. However, by using the Cholesky decomposition $F_\lm^{-1}=L_\lm\T  L_\lm$, it is possible to re-written the Fisher Kernel as an Euclidean dot-product, i.e. 
   % \begin{equation*}
   $ K(X,Y)=(\mathcal{G}_{\lm}^X)\T  \mathcal{G}_{\lm}^Y,$
  % \end{equation*}
    where
   % \begin{equation}\label{eq:fv}
    $\mathcal{G}_{\lm}^X=L_\lm{G}_{\lm}^X$ %$=L_\lm\nabla_\lm\log p(X|\lm)$
    %\end{equation}
    %The vector $\mathcal{G}_{\lm}^X$ 
     is the \emph{Fisher Vector} (FV) of $X$ \cite{perronnin10}. 
     
   Note that the FV is a fixed size vector whose dimensionality only depends on the dimensionality $m$ of the parameter $\lambda$.  
    The FV is further divided by $|X|$ in order to avoid the dependence on the sample size \cite{sanchez13} and $L_2$-normalized because, as proved in \cite{perronin10:improvingFK,sanchez13}, this is a way to cancel-out the fact that different images contain different amounts of image-specific information (e.g. the same object at different scales).

The distribution $p(\cdot|\lm)$, which models the generative process in the space of the data observation, can be chosen in various way.
  The Gaussian Mixture Model (GMM) is typically used to model the distribution of non-binary features %, like SIFT descriptors \cite{lowe99},
  considering that, as pointed in \cite{mclachlan2000}, any continuous
  distribution can be approximated arbitrarily well by an
  appropriate finite Gaussian mixture.
  %because one
  %can approximate with arbitrary precision any continuous distribution with a GMM \cite{titterington85,sanchez13} [\textit{ancora da
  %controllare cosa dice di preciso il riferimento \cite{titterington85}}].
  %
  % However, GMM could not catch the  inadequate (??)} to characterize binary feature, so () we choose $p$ to be a multivariate Bernoulli
  %Consequently, it is possible to choose a GMM also for modeling binary features. However,
  Since the Bernoulli distribution models an experiment that has
  only two possible outcomes (0 and 1), a reasonable alternative  to  characterize the distribution of a set of binary  features is to use a
  Bernoulli Mixture Model (BMM).

    \paragraph{FV and Binary Local Features}
 In this work we  derive and test an extension  of the FV built using BMM, called \textit{BMM-FV}, to encode binary features.  
    %Our approach follow the one used in \cite{sanchez13} for computing the traditional FV based on GMM.
%
%  In the rest of the paper we refer to the use of Fisher Vectors
%  with the Bernoulli Mixture model, to characterize binary local
%  features, as BMM-FV.
  Specifically, we chose $p(\cdot|\lm)$ to be multivariate Bernoulli mixture with
  $K$ components and parameters $\lm=\{w_k, \mu_{kd},\, k=1,\dots,
  K,\, d=1,\dots, D\}$:
  \begin{equation}
  p(x_t|\lm)=\sum_{k=1}^K w_k p_k(x_t)
  \end{equation}
  where
  \begin{equation}
  p_k(x_t)=\prod_{d=1}^D \mu_{kd}^{x_{td}}(1-\mu_{kd})^{1-x_{td}}
  \end{equation}
  and
  \begin{equation}\label{eq:constrain}
  \sum_{k=1}^K w_k=1, \quad w_k>0 \qquad \forall\,  k=1, \dots, K.
  \end{equation}

  To avoid enforcing explicitly the constraints in \eqref{eq:constrain}, we used the soft-max formalism \cite{krapac11,sanchez13} for the weight parameters:
  $
  w_k={\exp(\alpha_k)}/{\sum_{i=1}^K \exp(\alpha_i)}.
 $
 
 Given a set $X=\{ x_t,\, t=1,\dots, T\} $ of $D$-dimensional binary vectors $x_t\in \{0,1\}^D$ %, such as the binary local features extracted from an  image,  
 and assuming that the samples are independent 
%   %the the Fisher Vector of $X$ is given by
%   \begin{equation}
%   %\gl^X=\dfrac{1}{T} \sum_{t=1}^T L_\lm \nabla_\lm\log p(x_t|\lm).
%     \gl^X=\dfrac{1}{T}L_\lm\Glx=  \dfrac{1}{T} \sum_{t=1}^T L_\lm \nabla_\lm\log p(x_t|\lm).
%   \end{equation}
%   %by assuming that the samples are independent.
  we have that the score vector $\Glx$ with respect to the parameter $\lm=\{\alpha_k, \mu_{kd},\, k=1,\dots, K, \, d=1,\dots, D\}$
  is calculated (see %\ref{score vector}
  Appendix A) as the concatenation of
  \begin{align*}
  G_{\alpha_k}^X &=\sum_{t=1}^T \dfrac{\partial\log p(x_t|\lm)}{\partial \alpha_k}=\sum_{t=1}^T\left( \gamma_t(k)-w_k \right)\\
  G_{\mu_{kd}}^X &=\sum_{t=1}^T \dfrac{\partial\log p(x_t|\lm)}{\partial \mu_{kd}} =\sum_{t=1}^T
  \gamma_t(k)\left(\dfrac{x_{td}-\mu_{kd}}{\mu_{kd}(1-\mu_{kd})}\right)
  \end{align*}
  where $\gamma_t(k)=p(k| x_t,\lm)$ is the \emph{occupancy probability} (or posterior probability). The occupancy probability $\gamma_t(k)$ represents the probability for the observation $x_t$ to be generated by the $k$-th Bernoulli and it is calculated as
$
  \gamma_t(k)={w_k p_k(x_t)}/{\sum_{j=1}^K w_j p_j(x_t)}.
$
  
  The FV of $X$ is then obtained by normalizing the score $\Glx$ by the matrix $L_\lm$, which is the square root of the inverse of the FIM, and by the sample size $T$.  In the %\ref{FIM}
  Appendix B we provide an approximation of FIM under the assumption that the occupancy probability  $\gamma_t(k)$ is sharply peaked on a single value of $k$ for each descriptor $x_t$, obtained following an approach very similar to that used in \cite{sanchez13} for  the GMM case.
  %Please note that this approximation is different from that proposed in \cite{uchida13}.
  By using our FIM approximation, %and normalizing by the sample size $T$ (in order to avoid the dipedence of the sample)
  we got the following normalized gradient:
  \begin{align*}
  \mathcal G_{\alpha_k}^X &=\dfrac{1}{T\sqrt{w_k}}\sum_{t=1}^T \left(\gamma_t(k)-w_k\right) \\
  \mathcal G_{\mu_{kd}}^X %&=\dfrac{\sqrt{\mu_{kd}(1-\mu_{kd})}}{T\sqrt{w_k}}\sum_{t=1}^T \gamma_t(k)\left(\dfrac{x_{td}-\mu_{kd}}{{\mu_{kd}(1-\mu_{kd})}}\right)\nonumber\\
  &=\dfrac{1}{T\sqrt{w_k}}\sum_{t=1}^T \gamma_t(k)\left(\dfrac{x_{td}-\mu_{kd}}{\sqrt{\mu_{kd}(1-\mu_{kd})}}\right)
  \end{align*}
  
  The final BMM-FV is the concatenation of $\mathcal G_{\alpha_k}^X$ and  $\mathcal G_{\mu_{kd}}^X$ for $k=1,\dots, K$, $d=1,\dots, D$ and is
  therefore of dimension $K(D+1)$.
  
    \begin{table}[bt]
          \renewcommand{\arraystretch}{1.3}
          \small
          %% if using array.sty, it might be a good idea to tweak the value of
          %% \extrarowheight as needed to properly center the text within the cells
          \caption{Comparison of the structure of the FVs derived using BMM with that derived using GMM. Parameters for  BMM are  $\lm^B=\{w_k^B, \mu_{kd}^B,\, k=1,\dots, K,\, d=1,\dots, D\}$  and for GMM are $\lm^G=\{w_k^G, \mu_{kd}^G,\Sigma_k^G=\text{diag}(\sigma_{k1}^G,\dots,\sigma_{kD}^G,), \, k=1,\dots, K,\, d=1,\dots, D\}$,  where $w_{k}^B$, $\mu_{k}^B$ are the mixture weight and the mean vector of the $k$-th Bernoulli and $w_{k}^G$ $\mu_{k}^G$, $\Sigma_{k}^G$  are respectively the mixture weight, mean vector and covariance matrix of Gaussian $k$.}
          \label{tab:BMM.GMM} %to refer this table
          \centering
          \begin{tabular}{l}
               \noalign{\smallskip}\hline\noalign{\smallskip}
                      \multicolumn{1}{c}{\textbf{GMM-FV \cite{sanchez13}}}\\
                      \noalign{\smallskip}\hline\noalign{\smallskip}
                      $  \displaystyle
                      \mathcal G_{\alpha_k^G}^X =\frac{1}{T\sqrt{w_k^G}}\sum_{t=1}^T \left(\gamma_t^G(k)-w_k^G\right)
                      $\\
                      $\displaystyle
                      \mathcal G_{\mu_{kd}^G}^X =\frac{1}{T\sqrt{w_k^G}}\sum_{t=1}^T \gamma_t^G(k)\frac{x_{td}-\mu_{kd}^G}{{\sigma_{kd}^G}}
                      $ \\
                      $\displaystyle
                      \mathcal G_{\sigma_{kd}^G}^X =\frac{1}{T\sqrt{w_k^G}}\sum_{t=1}^T \gamma_t^G(k)\frac{1}{\sqrt{2}}\left[
                      \frac{(x_{td}-\mu_{kd}^G)^2}{(\sigma_{kd}^G)^2} -1
                      \right]
                      $\\
         
              \noalign{\smallskip}    \hline\noalign{\smallskip}
              \multicolumn{1}{c}{\textbf{BMM-FV} (our formalization)}  \\
              \noalign{\smallskip}\hline\noalign{\smallskip}
              $\displaystyle
              \mathcal G_{\alpha_k^B}^X =\frac{1}{T\sqrt{w_k^B}}\sum_{t=1}^T \left(\gamma_t^B(k)-w_k^B\right)
              $
              \\
              $\displaystyle
              \mathcal G_{\mu_{kd}^B}^X =\frac{1}{T\sqrt{w_k^B}}\sum_{t=1}^T \gamma_t^B(k)\frac{x_{td}-\mu_{kd}^B}{\sqrt{\mu_{kd}^B(1-\mu_{kd}^B)}}\qquad
              $   \\
               \noalign{\smallskip}    \hline\noalign{\smallskip}
                                  \multicolumn{1}{c}{\textbf{BMM-FV} (Uchida et. al \cite{uchida13})}  \\
                                  \noalign{\smallskip}\hline\noalign{\smallskip}
                                  ($\displaystyle
                                            \mathcal G_{\alpha_k^B}^X $ not explicitly derived in \cite{uchida13})
                                  \\
                                  $\displaystyle
                                  \mathcal G_{\mu_{kd}^B}^X =\dfrac{\sum_{t=1}^T \gamma_t(k)\frac{(-1)^{1-x_{td}}}{({\mu_{kd}^B})^{x_{td}}\left(1-\mu_{kd}^B\right)^{1-x_{td}}}}{T\sqrt{T w_k^B\left(\frac{\sum_{i=1}^K w_i^B\mu_{id}^B}{(\mu_{kd}^B)^2} +\frac{\sum_{i=1}^K w_i^B(1-\mu_{id}^B)}{(1-\mu_{kd}^B)^2}\right)}}      \qquad
                                  $   \\
                                  \noalign{\smallskip}\hline
          \end{tabular}
      \end{table}
      
   An extension of the FV by using the BMM has been also carried in \cite{uchida13,sanchez15}.
   Our approach differs from the one proposed in \cite{uchida13} in the approximation of the square root  of the inverse of the FIM (i.e., $L_\lm$) .  It is worth noting that our formalization preserves the structure of the traditional FV derived by using  the GMM, where Gaussian means and variances are replaced by Bernoulli means $\mu_{kd}$ and variances $\mu_{kd}(1-\mu_{kd})$ (see Table \ref{tab:BMM.GMM}).    
   
  In  \cite{sanchez15},  the FV formalism was generalized to a broaden family of distributions knows as\textit{ exponential family} that encompasses the Bernoulli distribution as well as the Gaussian one. However, \cite{sanchez15} lacks in an explicit definition of the FV and of the FIM approximation in the case of BMM  which was out of the scope of their work. % which gives a more general formulation.  
  Our formulation differs from that of \cite{sanchez15} in the choice of the parameters used in the gradient computation of the score function \footnote{A Bernoulli distribution $p(x)=\mu^x(1-\mu)^{1-x}$ of parameter $\mu$ can be written as exponential distribution $p(x)=exp(\eta x- log(1+e^\eta))$ where $\eta= \log(\frac{\mu}{1-\mu})$ is the \textit{natural parameter}.  In \cite{sanchez15} the score function is computed considering the gradient w.r.t. the {natural parameters} $\eta$ while in this paper we used the gradient w.r.t. the standard parameter $\mu$ of the Bernoulli (as also done in \cite{uchida13} ). }.  A similar difference holds  also for the FV computed on the GMM, given that  in \cite{sanchez15} the score function is computed w.r.t. the {natural parameters} of the Gaussian distribution rather than the mean and the variance  parameters which are typically used in literature for the FV representation
  \h{ }\cite{perronnin10,perronnin07,sanchez13}.  %In fact, the authors computed the gradient with respect the \textit{natural parameters} of an exponential distribution
Unfortunately, the authors of  \cite{sanchez15} didn't  experimentally compare the FVs obtained  using or not the natural parameters.
    
%   It is worth noting that our approach, differently from the one proposed in \cite{uchida13}, preserves the structure of the traditional FV derived by using  the GMM, where Gaussian means and variances are replaced by Bernoulli means $\mu_{kd}$ and variances $\mu_{kd}(1-\mu_{kd})$ (see Table \ref{tab:BMM.GMM}).
  S\`{a}nchez \cite{sanchez13} highlights that the FV derived from GMM can be computed in terms of the following  $0$-order and  $1$-order statistics: $ S_k^0=\sum_{t=1}^T \gamma_t(k) \, \in\R{}$, $S_k^1=\sum_{t=1}^T \gamma_t(k)x_t\, \in\R{D}$. %Our derivation of the FV, by using a BMM on binary features, could 
  Our BMM-FV can be also written in terms of these statistics as
%      \begin{align*}
%      S_k^0&=\sum_{t=1}^T \gamma_t(k) \, \in\R{} \\
%      S_k^1&=\sum_{t=1}^T \gamma_t(k)x_t\, \in\R{D}
%      \end{align*}
%      getting
      \begin{align*}
      \mathcal G_{\alpha_k}^X &=\dfrac{1}{T\sqrt{w_k}}(S_k^0-Tw_k) \\%T
      \mathcal G_{\mu_{kd}}^X &=\dfrac{S_{kd}^1 -\mu_{kd}S_k^0}{T\sqrt{w_k \mu_{kd}(1-\mu_{kd})}}.%T
      \end{align*}
  
  %S\`{a}nchez \cite{sanchez13} highlights that the FV derived from GMM could be computed in terms of  $0$-order and  $1$-order statistics. Our derivation of the FV, by using a BMM on binary features, could be also viewed in terms of the following $0$-order and  $1$-order statistics:
%    \begin{align*}
%    S_k^0&=\sum_{t=1}^T \gamma_t(k) \, \in\R{} \\
%    S_k^1&=\sum_{t=1}^T \gamma_t(k)x_t\, \in\R{D}
%    \end{align*}
%    getting
%    \begin{align*}
%    \mathcal G_{\alpha_k}^X &=\dfrac{1}{T\sqrt{w_k}}(S_k^0-Tw_k) \\%T
%    \mathcal G_{\mu_{kd}}^X &=\dfrac{S_{kd}^1 -\mu_{kd}S_k^0}{T\sqrt{w_k \mu_{kd}(1-\mu_{kd})}}.%T
%    \end{align*}
%    
    We finally used power-law and $L_2$ normalization to improve the effectiveness of the BMM-FV approach.

 \subsection{Combination of \h{Convolutional Neural Network Features and Aggregations of Binary Local Feature}}\label{sec:cnn}
Convolutional \h{N}eural \h{N}etworks (CNNs) \cite{LeCun2015} have brought breakthroughs in the computer vision area by improving the state-of-the-art in several domains, such as image retrieval, image classification, object recognition, and action recognition.
Depp CNN allows a machine to automatically learn representations of data with multiple levels of abstraction which can be used for detection or classification tasks. CNNs are neural networks specialized for data that has a grid-like topology as image data. The applied discrete convolution operation results in a multiplication by a matrix which has several entries constrained to be equal to other entries. Three important ideas are behind the success CNNs: sparse connectivity, parameter sharing, and equivariant representations \cite{DeepLearning2016}.

In image retrieval, the activations produced by an image within the top layers of the CNN have been successfully used as a high-level descriptors of the visual content of the image \cite{DeCaf}.
%In \cite{razavian2014} the same approach was adopted for evaluating the CNN representation in a visual instance retrieval task. 
The results reported in \cite{razavian2014cnn} shows that %the activations produced within the top layers of the CNN, 
these CNN features, compared by using the Euclidean distance, achieve state-of-the-art quality in terms of mAP.
Most of the papers reporting results obtained using the CNN features %obtained from a 
maintain the Rectified Linear Unit (ReLU) transform \cite{DeCaf,razavian2014cnn,chandrasekhar2015}, i.e., negative activations values are discarded replacing them with 0.
%In our experiments, we also reported the results obtained without the RELU as in \cite{babenko2014neural}.
%In fact, while the RELU, being a non-linear operation, has been proved to be very effective as activation, the negative values discarded by using this operation could be also exploited in the visual feature.
Values are typically $L_2$ normalized \cite{babenko2014neural,razavian2014cnn,chandrasekhar2015} and we did the same in this work.
In Section \ref{sec:expsetting} we describe the CNN model used in our experiments.

 Recently, in \cite{chandrasekhar2015} it has been shown that the information provided by the FV built upon SIFT helps to further improve the retrieval performance of the CNN features and a combination of FV and CNN features has been used as well \cite{chandrasekhar2015,amato16:JOCCH}. However, the benefits of such combinations are clouded by the cost of extracting SIFTs that can be considered to high with respect to the cost of computing the CNN features (see Table \ref{tab:efficiencyComparison}).
  % retrieval performance of CNN features has been improved thanks to their combination with FVs built upon SIFTs. However, the benefit provided by the combination is clouded by the costly SIFT extraction.
Since the extraction of binary local features is up two times faster than SIFT, in this work we also investigate the combination of CNN features with 
\h{aggregations of binary local feature, including BMM-FV.}
%the BMM-FV built upon binary local features. 

 We combined BMM-FV and CNN using the following approach. Each image was represented by a couple $(c, f)$, where $c$ and $f$ were respectively the CNN descriptor and the BMM-FV of the image. Then, we evaluated the distance $d$ between two couples $(c_1, f_1)$ and $(c_2, f_2)$  as the convex combination between the $L_2$ distances of the CNN descriptors (i.e. $\|c_1-c_2\|_2$) and the BMM-FV descriptors (i.e. $\|f_1-f_2\|_2$). In other words, we defined the distance between two couples $(c_1, f_1)$ and $(c_2, f_2)$ as
   \begin{equation}\label{eq:combination}
   d\big((c_1, f_1), (c_2, f_2)\big)= \alpha\,\|c_1-c_2\|_2 +(1-\alpha)\,\|f_1-f_2\|_2
   \end{equation}
   with $0\leq \alpha\leq1$. Choosing $\alpha=0$ corresponds to use only FV approach, while $\alpha=1$ correspond to use only CNN features.
   Please note that in our case both the FV and the CNN features are $L_2$ normalized \h{so the distance function between the CNN descriptors has the same range value of the distance function between the BMM-FV descriptors.} 
   
   \h{
   Similarly, combinations between CNN features and other image descriptors, such as GMM-FV, VLAD, and BoW can be considered by using the convex combination of the respective distances. Please note that whenever the range of the two used distances is not the same, the distances should be rescaled before the convex combination  (e.g. divide each distance function by its maximum value).}
   
   \begin{table*}[bt]\footnotesize
 \renewcommand{\arraystretch}{1}
 	\caption{Average time costs for computing various image representations %, namely CNN feature, FV encoding and SIFT/ORB local features extraction 
 		using a CPU implementation. The cost of computing the CNN feature of an image was estimated using \h{a} pre-learned AlexNet model and the Caffe framework \cite{caffe2014} \h{on an Intel i7 3.5 GHz}. The values related to the FV refers only to the cost of aggregating the local descriptors of an image into a single vector and do not encompass the cost of extracting the local features, neither the learning of the Gaussian or the Bernoulli Mixture Model which is calculated off-line. The cost of computing FV varies \h{proportionally with $TKD$, where $T$ is the number of local features extracted from an image, $K$ is the number of mixtures of Gaussian/Bernoulli, and $D$ is the dimensionality of each local feature}; we reported the approximate cost for \h{$T=2,000$ and $KD=64*64$ and $KD=64*256$ on an Intel i7 3.5 GHz}. The cost of SIFT/ORB local feature extraction was estimated according to \cite{heinly12} by considering about $2,000$ features per image.}\label{tab:efficiencyComparison}
 	\centering
 	\arrayrulecolor{gray}
 	\setlength\tabcolsep{2pt} 
 	%Performance comparison of BoW, VLAD and FV aggregations of ORB binary features
 	\begin{tabular}{@{}
 			>{\arraybackslash} p{0.25\textwidth}| %method
 			>{\centering\arraybackslash} p{0.15\textwidth}| 
 			>{\centering\arraybackslash} p{0.25\textwidth} |
 			>{\centering\arraybackslash} p{0.15\textwidth} |
 			>{\centering\arraybackslash} p{0.15\textwidth} 
 			@{}}
 		%	\toprule[1.1pt]
 		%\hline  %\toprule
 		& CNN  & {FV} Encoding & SIFT  & ORB\\
 		%		   	& &(encoding only)&&\\
 		%	\midrule[1.1pt]	
 		\hline	
 %		\multirow{2}{0.28\textwidth}{\centering Run time (CPU) ms/ image $\sim$}&
 	\multirow{2}{0.2\textwidth}{\centering Computing time \\ per image }&
 		\multirow{2}{0.15\textwidth}{\centering ${\sim}300$ ms}&
 		${\sim}40$ ms [\h{$KD=64*64$}]&
 		\multirow{2}{0.15\textwidth}{\centering ${\sim}1200$ ms}& \multirow{2}{0.1\textwidth}{\centering ${\sim}26$ ms}\\
 		& & ${\sim}160$ ms [\h{$KD=256*64$}]& &\\
 		%	\bottomrule[1.1pt]
 		%\hline
 	\end{tabular}
 \end{table*} 

%\subsection{Combination of FV and CNN Features}\label{sec:combination}
%  \todoinL{Lunedi-martedi}
 \section{Experiments} \label{sec:exp}
 %%
 %%PAST SIMPLE: what we did + how we did
 %%          what happened during our experiments
 %%SIMPLE PRESENT:
 %%              -what others have done
 %%              -rif table anf figure
 %%              -To say what the implications are.Typically after show,explain, highlight,believe, mean
 In this section we evaluate and compare the performance of the
 techniques described in this paper to aggregate binary local
 descriptors. Specifically, in the Subsection \ref{sec:expCompEncoding} we compare the BoW, the VLAD, the FV based on the GMM,
 and the BMM-FV approach to aggregate ORB binary features. 
 %We evaluate the performance on two standard benchmarks and we also compare the results against matching the local features without any aggregation. 
%  We will see that some approaches, i.e. VLAD and BMM-FV, provide very compact image representations with retrieval  performance comparable to that obtained with the direct matching of the binary features.
  Since the BMM-FV achieved the best results over the other tested approaches, in the Subsection \ref{sec:expCompFVandCNN} we further evaluate the performance of the BMM-FVs using different binary features (ORB, LATCH, AKAZE) and combining them  with the CNN features. \h{Finally, in the Subsection} \ref{sec:largescale}\h{, we report experimental results on large scale.}

%The experiments were performed on two standard benchmark for image retrieval ()

  In the following, we first introduce the datasets used in the evaluations (Subsection \ref{sec:datasets}) and we describe our experimental setup (Subsection \ref{sec:expsetting}). We then report results and their analysis.
 
 \subsection{Datasets} \label{sec:datasets}
The  experiments  were  conducted  using  two benchmark datasets, namely \textit{INRIA Holidays}  \cite{jegou08} and \textit{Oxford5k }\cite{philbin07}, that are publicly available and often used in the context of  image retrieval \cite{jegou10:VLAD,zhao13,jegou08,arandjelovic12:rootsift,perronnin10,jegou12,tolias14}.

INRIA Holidays \cite{jegou08}
is a collection of $1,491$ images which mainly contains personal holidays photos. The images are of high resolution and represent a large variety of scene type (natural, man-made, water, fire effects,
etc). The dataset contains 500 queries, each of which represents a distinct scene or object. For each query a list of positive results is provided.  As done by the authors of the dataset,  we resized the images to a maximum of $786,432$ pixels ($768$ pixels for the smaller dimension) before \h{extracting} the local descriptors.

Oxford5k \cite{philbin07}
consists of $5,062$ images collected from Flickr. The dataset comprise $11$ distinct Oxford buildings together with distractors.
There are $55$ query images: $5$ queries for each building. %defined by a rectangular region delimiting a building on an image.
The collection  is provided with a comprehensive ground truth. For each query there are four image sets: \emph{Good} (clear pictures of the object represented in the query), \emph{OK}  (images where more that $25\%$ of the object is clearly visible),  \emph{Bad} (images where the object is not present) and \emph{Junk} (images where less than $25\%$ of the object is visible or images with high level of distortion).

As in many other articles, e.g. \cite{jegou10:VLAD,jegou08,philbin08,jegou12}, all the learning stages (clustering, etc.) were performed off-line using independent image collections.
%{Flickr60k} dataset \cite{jegou08} was used as training set for the experiments on INRIA Holidays, while {Paris6k} dataset \cite{philbin08} was used for the experiments on Oxford5k.
\emph{Flickr60k} dataset \cite{jegou08} was used as training set for INRIA Holidays. It is composed of $67,714$ images \h{randomly extracted}  from Flickr.
% {Flickr60k} is composed of $67,714$ images extracted randomly from Flickr. Compared to INRIA Holidays, the Flickr dataset is slightly biased, because it includes low-resolution images and more photos of humans.
The experiments on Oxford5k were conducted performing the learning stages on \emph{Paris6k} dataset \cite{philbin08}, that contains $6,300$ high resolution images obtained from Flickr by searching for famous Paris landmarks.

\h{For large-scale experiments we combined the Holidays dataset with the 1 million MIRFlickr dataset} \cite{huiskes08},\h{ used as distractor set as also done in}  \cite{jegou08,Amato2016}. \h{
Compared to Holidays, the Flickr datasets is slightly biased, because it includes low-resolution images and more photos of humans.}
\subsection{Experimental settings} \label{sec:expsetting}
In the following we report some details on how the features for the various approaches were extracted.
%Moreover, we describe how we combined FV and CNN features.\\
\begin{description}
\item[\textbf{Local features.}]
In the experiments we used ORB \cite{rublee11}, LATCH \cite{levi15_LATCH}, and AKAZE \cite{akaze} binary local features that \h{were} extracted by using OpenCV (Open Source Computer Vision Library)\footnote{\url{http://opencv.org/}}.
We detected up to $2,000$ local features per image. 
%Since the INRIA Holidays photos are of high resolutions we resized the images to a maximum of $786,432$ pixels ($768$ pixels for the smaller dimension) before computing the descriptors, following the indication provided by the authors of the dataset.

\item[\textbf{Visual Vocabularies and Bernoulli/Gaussian Mixture Models.}] The visual vocabularies used for building the BoW and VLAD representations were computed using several clustering algorithms, i.e. $k$-medoids, $k$-majority and $k$-means. %, on the training sets (namely Flickr60k for INRIA Holidays and  Paris6k for Oxford5k). 
The $k$-means algorithm was applied to the binary features by treating the binary vectors as real-valued vectors. 
%%%All the learning stages, i.e. $k$-means, $k$-medoids, $k$-majority and the estimation of GMM/BMM, were performed using in order of $10^6$ descriptors randomly selected from all the ORBs extracted on the training set (namely Flickr60k for INRIA Holidays and  Paris6k for Oxford5k).
%\textbf{Bernoulli and Gaussian Mixture Models:} 
%The Bernoulli/Gaussian Mixture Models and the Fisher Vector representations were computed by using our Visual Information Retrieval library that is publicly available on GitHub\footnote{https://github.com/ffalchi/it.cnr.isti.vir}. 
%The Bernoulli and the Gaussian Mixture Models were computed using our Visual Information Retrieval library that is publicly available on GitHub\footnote{https://github.com/ffalchi/it.cnr.isti.vir}. 
The parameters
$\lambda^B=\big\{w_k^B, \mu_{kd}^B\big\}_{k=1,\dots, K,\,
    d=1,\dots, D}$ of the BMM and $\lambda^G=\big\{w_k^G,\mu_{kd}^G,\sigma_{kd}^G\big\}_{k=1,\dots, K,\, d=1,\dots, D}$ of the
GMM (where $K$ is the number of mixture components and $D$ is the dimension of each local descriptor) were learned independently by optimizing a maximum-likelihood
criterion with the Expectation Maximization (EM) algorithm
\cite{bishop06}. 
EM is an iterative method that is deemed to have
converged when the change in the likelihood function, or
alternatively in the parameters $\lambda$, falls below some
threshold $\epsilon$.
As stopping criterion we used the convergence in $L_2$-norm of the mean
parameters, choosing $\epsilon=0.05$.
As suggested in \cite{bishop06}, the BMM/GMM parameters  used in EM algorithm were initialized with: (a) $1/K$ for the mixing coefficients $w_{k}^B$ and $w_{k}^G$; (b) random values chosen uniformly in the range $(0.25, 0.75)$, for the BMM means $\mu_{kd}^B$; (c) centroids precomputed using $k$-means for the GMM means  $\mu_{kd}^G$; (d) mean variance of the clusters found using $k$-means for the diagonal elements $\sigma_{kd}^G$ of the GMM covariance matrices.
% $\lm^B=\{w_k^B, \mu_{kd}^B,\, k=1,\dots, K,\, d=1,\dots, D\}$
%$\lm^G=\{w_k^G, \mu_{kd}^G,\Sigma_k^G=\text{diag}(\sigma_{k1}^G,\dots,\sigma_{kd}^G,), \,$

All the learning stages, i.e. $k$-means, $k$-medoids, $k$-majority and the estimation of GMM/BMM, were performed using in order of 1M descriptors randomly selected from the local features extracted on the training sets (namely Flickr60k for INRIA Holidays and  Paris6k for Oxford5k).

\item[\textbf{BoW, VLAD, FV.}]
The various encodings of the local feature (as well as the visual vocabularies and the BMM/GMM) were computed by using our Visual Information Retrieval library that is publicly available on GitHub\footnote{https://github.com/ffalchi/it.cnr.isti.vir}. 
These representations are all parametrized by a single integer $K$. It corresponds to the number of centroids (visual words) used in BoW and VLAD, and to the number of mixture components of GMM/BMM used in FV representations.

For the FVs, we used only the components
$\mathcal G_{\mu}$ associated with the mean vectors because,
as happened in the non-binary case, we observed that the components
related to the mixture weights do not improve the results.

As a common post-processing step \cite{perronin10:improvingFK,jegou12}, both the FVs and the VLADs were power-law normalized and subsequently $L_2$-normalized. The power-law normalization is parametrized by a constant $\beta$ and it is defined as $x\to |x|^\beta \text{sign}(x)$. In our experiments we used $\beta=0.5$.\\
We also applied PCA to reduce VLAD and FV dimensionality. The projection matrices were estimated on the training datasets.

\item[\textbf{CNN features.}] 
We used the pre-trained HybridNet \cite{zhou2014} model, downloaded from the Caffe Model Zoo\footnote{\url{https://github.com/BVLC/caffe/wiki/Model-Zoo}}. The architecture of  HybridNet is the same as the BVLC Reference CaffeNet\footnote{\url{https://github.com/BVLC/caffe/tree/master/models/bvlc_reference_caffenet}} which mimics the original AlexNet \cite{krizhevsky2012}, with minor variations as described in \cite{caffe2014}.  It has $8$ weight layers ($5$ convolutional + $3$ fully-connected). The model has been trained on $1,183$ categories ($205$ scene categories from Places Database \cite{zhou2014} and $978$ object categories from ImageNet \cite{deng2009}) with about $3.6$ million images. 
%The input image are fixed-size to $227\times 227$ RGB. 
%The architecture takes a $227\times 227$ RGB image as input.

In the test phase we used Caffe and we extracted the output of the first fully-connected layer (\textit{fc6}) after applying the Rectified Linear Unit (\textit{ReLU}) transform. The resulting $4,096$-dimensional descriptors were $L_2$ normalized.

As preprocessing step we warped the input images to the canonical resolution of $227\times 227$ RGB (as also done in \cite{DeCaf}).
\item[\textbf{Feature comparison and performance measure.}] The cosine similarity in conjunction with a term weighting scheme (e.g., tf-idf) is adopted for evaluating the similarity between BoW representations, while the Euclidean distance is used to compare VLAD, FV and CNN-based image signatures.\h{ Please note that the Euclidean distance is equivalent to the cosine similarity whenever the vectors are $L_2$-normalized, as in our case}\footnote{\h{
To search a database for the objects similar to a query we can use either a \textit{similarity} function or a \textit{distance} function. In the first case, we search for the objects with greatest similarity to the query. In the latter case, we search for the objects with lowest distance from the query. A similarity function is said to be equivalent to a distance function if the ranked list of the results to query is the same.
For example, the Euclidean distance between two vectors ($\ell_2(x_1,x_2)={\|x_1-x_2\|}_2$) is equivalent to the cosine similarity ($s_{\text{cos}}(x_1,x_2)=(x_1\cdot x_2)/({\|x_1\|}_2{\|x_2\|}_2)$) whenever the vectors are $L_2$- normalized (i.e. $\|x_1\|_2=\|x_2\|_2=1$). In fact, in such case, $s_{\text{cos}}(x_1,x_2)=1-\frac{1}{2}{\ell_2(x_1,x_2)}^2$, which implies that the ranked list of the results to a query is the same (i.e., $\ell_2(x_1,x_2)\le \ell_2(x_1,x_3)$ iff $s_{\text{cos}}(x_1,x_2)\geq s_{\text{cos}}(x_1,x_3)\, \forall\, x_1,x_2,x_3 $).}}.

The image comparison based on the \textit{direct matching} of the local features \h{(}i.e. without aggregation) was performed adopting the
distance ratio criterion proposed in \cite{lowe04,heinly12}.
Specifically, candidate matches to local features of the image
query are identified by finding their nearest neighbors in the
database of images. Matches are discarded if the ratio of the
distances between the two closest neighbors is above the 0.8
threshold. Similarity between two images is computed as the
percentage of matching pairs with respect to the total local
features in the query image.

The retrieval performance of each method was measured by the \textit{mean average precision} (mAP). In the experiments on INRIA Holidays, we computed the average precision after removing the query image from the ranking list.
In the experiments on Oxford5k, we removed the \textit{junk} images from the ranking before computing the average precision, as recommended in  \cite{philbin07} and in the evaluation package provided with the dataset.

\end{description}

%\textbf{Combination of FV and CNN features:}  {FVs and outputs of intermediate layers of CNN have complementary behavior under some image transformations. }
%In fact, the FVs (computed from SIFT or SIFTPCA) are robust to image rotation while the CNN features have limited level of rotation invariance. Additionally, in \cite{chandrasekhar2015} extensive experiments on benchmark dataset for image retrieval have showed that CNN features generally are less affected by small scale changes than FV.
%In order to leverage the positive aspects of both these methods, in \cite{chandrasekhar2015} a fusion of FV and CNN features has been proposed.
%
%In this paper, we evaluated the combination of FV and CNN features using the following approach. Each image was represented by a couple $(c, f)$, where $c$ and $f$ were respectively the CNN descriptor and the FV descriptor of the image. Then, we evaluated the distance $d$ between two couples $(c_1, f_1)$ and $(c_2, f_2)$  as the convex combination of the $L_2$ distances of the CNN descriptors (i.e. $\|c_1-c_2\|_2$) and the FV descriptors (i.e. $\|f_1-f_2\|_2$). In other words we defined
%\begin{equation}\label{eq:comb}
%d\big((c_1, f_1), (c_2, f_2)\big)= \alpha\,\|c_1-c_2\|_2 +(1-\alpha)\,\|f_1-f_2\|_2
%\end{equation}
%with $0\leq \alpha\leq1$. Choosing $\alpha=0$ corresponds to use only FV approach, while $\alpha=1$ correspond{s} to use only CNN features.
  
% \subsection{Results}
% \label{sec:res}
%  \todoinL{martedi}
 
\subsection{Comparison of Various Encodings of Binary Local Features}\label{sec:expCompEncoding}
  In Table \ref{tab:tab1} we summarize the retrieval performance of
    various aggregation methods applied to ORB features, i.e. the BoW, the VLAD, the FV based on the GMM,
       and the BMM-FV.  In addition, in the last line of the table we reports the results obtained without
         any aggregation, that we refer to as the \emph{direct matching} of local features, which was performed adopting the 
           distance ratio criterion as previously described in the Subsection \ref{sec:expsetting}.
  
%  In this section we evaluate and compare the performance of the
%   techniques described in this paper to aggregate binary local
%   descriptors. Specifically in the Subsection \ref{sec:expCompEncoding} we compare the BoW, the VLAD, the FV based on the GMM,
%   and the BMM-FV approach to aggregate ORB binary features. We evaluate the performance on two standard benchmarks and we also compare the results against matching the local features without any aggregation. 
   In our experiments the FV derived as in \cite{uchida13} obtained very similar performance to that of our BMM-FV, thus we have reported just the results obtained by using our formulation. 
   Furthermore, we have not experimentally evaluated the FVs computed using the gradient with respect to the \textit{natural parameters} of a BMM or a GMM as described in \cite{sanchez15}, because the evaluation of the retrieval performance obtained using or not the natural parameters in the derivation of the score function is a more general topic which reserve to be further investigated outside the specific context of the encodings binary local features.

  %Table \ref{tab:tab1} summarizes the retrieval performance of   various aggregation methods applied to ORB features. 
%  In addition,
%  the last line of the table also reports the results obtained without
%  any aggregation. We refer to the technique that does not use
%  aggregation, as the \emph{direct matching} of local features.
%For
%  direct matching, image comparison is performed adopting the
%  distance ratio criterion proposed in \cite{lowe04,heinly12}.
%  Specifically, candidate matches to local features of the image
%  query, are identified by finding their nearest neighbors in the
%  database of images. Matches are discarded if the ratio of the
%  distances between the two closest neighbors is above the 0.8
%  threshold. Similarity between two images is computed as the
%  percentage of matching pairs with respect to the total local
%  features in the query image.
%  
%table
  
  %In the  context of image retrieval, the most common way of using
  %binary features is the brute-force matching strategy. In our
  %experiments, the mAP achieved using the direct matching of ORB
  %descriptors was \textbf{41.3\%} on INRIA Holiday and
  %\textbf{31.7\%} on Oxford5k. Since we are interested in
  %identifying which aggregation method is more suitable for binary
  %features, we summarize in Table \ref{tab:tab1} the retrieval
  %results obtained by various aggregations of ORB binary features.
  
      \begin{table}[tbp]
      \renewcommand{\arraystretch}{1}
      	\caption{Performance evaluation of various aggregation methods applied on ORB binary features. \textit{$K$} indicates the number of centroids (visual words) used in BoW and VLAD and the number of mixture components of GMM/BMM used in FV; \textit{dim} is the number of components of each vector representation.\h{Bold numbers denote maxima in the respective column.}
      	}\label{tab:tab1}
      	%   \label{table_example}
      	\centering
      		\arrayrulecolor{gray}
      	\scriptsize %footnotesize	\footnotesize
      	%Performance comparison of BoW, VLAD and FV aggregations of ORB binary features
      	\begin{tabular}{
      		>{\centering\arraybackslash}	p{0.1\columnwidth}
      				>{\centering\arraybackslash}p{0.17\columnwidth}
      			>{\centering\arraybackslash}p{0.13\columnwidth}
      			>{\centering\arraybackslash}p{0.06\columnwidth}
      			>{\raggedleft\arraybackslash}p{0.12\columnwidth} 
                >{\centering\arraybackslash}p{0.087\columnwidth}
      			>{\centering\arraybackslash}p{0.087\columnwidth}@{}}\toprule[1.1pt]
      		   
      		\multirow{2}{0.1\columnwidth}{\centering\textbf{Method}} &   \multirow{2}{0.17\columnwidth}{\centering\textbf{Local Feature}} &   \multirow{2}{0.13\columnwidth}{\centering\textbf{Learning method}}
      		&   \multirow{2}{0.06\columnwidth}{\centering{$\mathbf{K}$}}  &
      		\multirow{2}{0.12\columnwidth}{\centering{\textbf{dim}}}  &   \multicolumn{2}{c}{\centering\textbf{mAP}}          \\
      		&       &           &   &   & \multirow{1}{0.087\columnwidth}{\scriptsize Holidays} &   \multirow{1}{0.087\columnwidth}{\scriptsize Oxford5k}   \\
      		\midrule[1.1pt]
	      	\scriptsize BoW     &   \scriptsize ORB &   $k$-means  &    20,000    &    20,000    &    44.9   &     22.2 \\
      		\scriptsize BoW     &   \scriptsize ORB &   $k$-majority  &    20,000    &    20,000    & 44.2      &     22.8 \\
      		\scriptsize BoW     &   \scriptsize ORB &   $k$-medoids   &    20,000    &    20,000    & 37.9 		      &      18.8 \\
      	\midrule
      	\scriptsize VLAD    &   \scriptsize ORB & {\centering $k$-means} &   64  &   16,384 &  47.8    &      23.6 \\ % 	>{\centering\arraybackslash}
			      	      	 &  	 &    &    &{\tiny PCA$\rightarrow$}\,1,024 & 46.0      	      		     &      23.2 \\
      	      		    &   &    &     &{\tiny PCA$\rightarrow$}\,\,\,\,\,128 &  30.9		     &      19.3\\
      			\scriptsize VLAD    &   \scriptsize ORB &   $k$-majority  &   64  &   16,384 &    32.4		   &       16.6\\
      			\scriptsize VLAD    &   \scriptsize ORB &   $k$-medoids   &   64  &   16,384 &  30.6     &       15.6\\

      	\midrule
      		\scriptsize FV  &   \scriptsize ORB &   \scriptsize GMM &   64  &   16,384 &   42.0   &       20.4\\
      		\scriptsize       &    &    &   &{\tiny PCA$\rightarrow$}\,1,024    &    42.6		   &      20.3 \\
      		\scriptsize   &   &    &    &{\tiny PCA$\rightarrow$}\,\,\,\,\,128    &   35.5   &     19.6  \\
      	\midrule
      		\scriptsize FV  &    &   \scriptsize BMM &   64  &   16,384 &   49.6		   &   \textbf{24.3}   \\
      		\scriptsize   &    &   &    &{\tiny PCA$\rightarrow$}\,1,024 &   \textbf{51.3}   &     23.4  \\
      		\scriptsize   &    &    &    &{\tiny PCA$\rightarrow$}\,\,\,\,\,128 &  44.6   &    19.1   \\
      	\midrule
      		\scriptsize No-aggr.  &   \scriptsize ORB &  \-- &   \--  &  &   38.1   &  31.7    \\
      		\bottomrule[1.1pt]
      	\end{tabular}
      \end{table}

   \begin{table}[tbp]
    	\renewcommand{\arraystretch}{1}
    	\caption{Aggregation methods on non-binary local features. Results are reported from  \cite{jegou10:VLAD,jegou12}.
    	}\label{tab:res-literature}
    	%   \label{table_example}
    	\centering
    		\arrayrulecolor{gray}
	\scriptsize %footnotesize
    	{%State-of-the-art performance comparison of BoW, VLAD and FV aggregations of SIFT features
    		\begin{tabular}{
    				>{\centering\arraybackslash}p{0.1\columnwidth}
    				p{0.17\columnwidth}
    				>{\centering\arraybackslash}p{0.12\columnwidth}
    				>{\centering\arraybackslash}p{0.06\columnwidth}
    				>{\raggedleft\arraybackslash}p{0.12\columnwidth}
    				>{\centering\arraybackslash}p{0.087\columnwidth}
    				>{\centering\arraybackslash}p{0.087\columnwidth}@{}}\toprule[1.1pt]
    			\multirow{2}{0.1\columnwidth}{\centering\textbf{Method}} &   \multirow{2}{0.17\columnwidth}{\textbf{Local Feature}} &   \multirow{2}{0.12\columnwidth}{\centering\textbf{Learning method}}
    			&   \multirow{2}{0.06\columnwidth}{\centering{$\mathbf{K}$}}  &
    			\multirow{2}{0.12\columnwidth}{\centering{\textbf{dim}}}  &   \multicolumn{2}{c}{\centering\textbf{mAP}}          \\
    			&       &           &   &   & \multirow{1}{0.087\columnwidth}{\scriptsize Holidays} &   \multirow{1}{0.087\columnwidth}{\scriptsize Oxford5k}   \\
    				\midrule[1.1pt]
    			\scriptsize  BoW     &    \scriptsize SIFT    &   $k$-means & 20,000  & 20,000  &   40.4   &   -   \\
    			\scriptsize BoW   &  {\scriptsize{SIFT}\,PCA\,64}    &   $k$-means & 20,000 & 20,000 &   43.7    &   35.4    \\
    		%	\scriptsize BoW  &  {\scriptsize{SIFT}\,PCA\,64}   &   $k$-means & 20\,000 &   {\tiny PCA$\rightarrow$}\,128 &   45.2    &   19.4    \\  
    	  			\midrule
    			\scriptsize VLAD     & \scriptsize   SIFT    &   $k$-means &   64  &       8,192   &   52.6    &   -   \\
    		%	\scriptsize VLAD     &  \scriptsize SIFT &   $k$-means &   64  &   {\tiny PCA$\rightarrow$}\,128 &   51.0    &   -   \\
    			   &   &   &     &   {\tiny PCA$\rightarrow$}\,128 &   51.0    &   -   \\
    			\scriptsize VLAD      & {\scriptsize{SIFT}\,PCA\,64}     &   $k$-means &   64  &           4,096  &   55.6    &   37.8    \\
    		%	\scriptsize VLAD      &  {\scriptsize{SIFT}\,PCA\,64}     &   $k$-means &   64  &   {\tiny PCA$\rightarrow$}\,128 &   55.7    &   28.7    \\ 
    		    &      &    &     &   {\tiny PCA$\rightarrow$}\,128 &   55.7    &   28.7    \\
    		 	\midrule
    			\scriptsize FV   & \scriptsize SIFT    &   \scriptsize GMM &   64  &           8,192   &   49.5    &   -   \\
    		%	\scriptsize FV   & \scriptsize SIFT    &   \scriptsize GMM &   64  &   {\tiny PCA$\rightarrow$}\,128 &   49.2    &   -   \\
    		  &     &    &     &   {\tiny PCA$\rightarrow$}\,128 &   49.2    &   -   \\
    			\scriptsize FV   &  {\scriptsize{SIFT}\,PCA\,64}    &  \scriptsize GMM &   64  &           4,096  &   59.5    &   41.8    \\
    		%	\scriptsize FV   &  {\scriptsize{SIFT}\,PCA\,64}   &   \scriptsize GMM &   64  &   {\tiny PCA$\rightarrow$}\,128 &   56.5    &   30.1   \\
    		  &    &    &     &   {\tiny PCA$\rightarrow$}\,128 &   56.5    &   30.1   \\
    			\bottomrule[1.1pt]
    		\end{tabular}}
    	\end{table}

  Among the various baseline aggregation methods (i.e. without using PCA), the
  {BMM-FV} approach achieves the best retrieval performance, that is
  a mAP of \emph{49.6\%} on Holidays and \textbf{24.3\% }on
  Oxford. PCA dimensionality reduction from $16,384$ to $1,024$
  components, applied on BMM-FV, marginally reduces the mAP on
  Oxford5k, while on Holiday allows us to get \textbf{51.3\%} that is,
  for this dataset, the best result achieved between all the other
  aggregation techniques tested on ORB binary features.
  %The PCA reduction to only 128 components, instead, give a slight loss in performance than using the full BMM-FV ($-1.5$ \% on Holiday and $-5.2 \% $ on Oxford5k).
  
  Good results are also achieved using VLAD in conjunction with
  $k$-means, which obtains a mAP of \emph{47.8\%} on Holidays and
  \emph{23.6\%} on Oxford5k. 
  %he use of PCA, also in this case,  improves the results on Holidays, reaching \textbf{45.7\%} when  moving from $16\, 384$ to $1\,024$ components. 
 
  The BOW representation allows to get a mAP of
  \emph{44.9\%}/\emph{44.2\%}/\emph{37.9\%} on Holidays and
  \emph{22.2\%}/\emph{22.8\%}/\emph{18.8\%} on Oxford5k using
  respectively $k$-means/$k$-majority/$k$-medoids for the learning
  of a visual vocabulary of $20,000$ visual words.
  
  The GMM-FV method gives results slight worse than BoW: \emph{42.0\%}
  of mAP on Holidays and \emph{20.4\%} of mAP on Oxford5k. The use
  of PCA to reduce dimensions from $16,384$ to $1,024$ lefts
  the results of GMM-FV  on Oxford5k substantially unchanged while slightly 
  improved the mAP on Holidays (\emph{42.6\%}).
  
  Finally, the worst performance are that of VLAD in combination with
  vocabularies learned by $k$-majority (\emph{32.4\%} on Holidays
  and \emph{16.6\%} on Oxford) and $k$-medoids (\emph{30.6\%} on
  Holidays and \emph{15.6\%} on Oxford).
  
  It is generally interesting to note that on INRIA Holidays, the 
  VLAD with $k$-means, the BoW with  $k$-means/$k$-majority, and the FVs are better than direct match. In
  fact, mAP of direct matching of ORB descriptors is
  \emph{38.1\%} while on Oxford5k the direct matching %is slighly betterreaching
  reached a mAP of \emph{31.7\%}.

      \begin{table}[tbp]
      \renewcommand{\arraystretch}{1}
      	\caption{Retrieval performance of our BMM-FV on INRIA Holidays and Oxford5k. \textit{$K$} is the number of BMM mixtures. \textit{dim} is the number of components of the final vector representation. \h{Bold numbers denote maxima in the respective column.} }\label{tab3}
      	\centering
      		\arrayrulecolor{gray}
      		\scriptsize %footnotesize
      	\subtable[Performance evaluation for increasing number $K$ of Bernullian mixture components\label{tab:fvBmm}]{
      		\begin{tabular}{  >{\centering\arraybackslash}p{0.06\columnwidth} >{\raggedleft\arraybackslash}p{0.175\columnwidth}| >{\centering\arraybackslash}p{0.15\columnwidth} >{\centering\arraybackslash}p{0.15\columnwidth}@{}}\toprule[1.1pt]
      			\multirow{2}{*}{\centering{$\mathbf{K}$}}  &
      			\multirow{2}{0.17\columnwidth}{\centering{\textbf{dim}}}  &   \multicolumn{2}{c}{\centering\textbf{mAP}}          \\
      			&               & \multirow{1}{*}{Holidays} &   \multirow{1}{*}{Oxford5k}   \\
      			\hline
      			4   &       1,024  &  32.0			     & 14.3     \\
      			8   &       2,048  &  38.2    &   17.4   \\
      			16  &       4,096  &  41.9    &   19.4   \\
      			32  &       8,192  &  45.9    &  21.3    \\
      			64  &       16,384 &  49.6    &  24.3    \\
      			128 &       32,768 &  52.3     &  26.4    \\
      			256 &       65,536 &  53.0     &   {27.3  } \\
      			512 &       131,072    &   \textbf{54.7}   &   \textbf{27.4}   \\
      	\bottomrule[1.1pt]
      		\end{tabular}}\\
      		\subtable[Results after dimensionality reduction when $K=64$ Bernoulli are used\label{tab:PCA}]{
      			\begin{tabular}{ >{\centering\arraybackslash}p{0.06\columnwidth} >{\raggedleft\arraybackslash}p{0.175\columnwidth}| >{\centering\arraybackslash}p{0.15\columnwidth} >{\centering\arraybackslash}p{0.15\columnwidth}@{}}\toprule[1.1pt]
      				\multirow{2}{*}{\centering{$\mathbf{K}$}}  &
      				\multirow{2}{0.17\columnwidth}{\centering{\textbf{dim}}}  &   \multicolumn{2}{c}{\centering\textbf{mAP}}          \\
      				&               & \multirow{1}{*}{Holidays} &   \multirow{1}{*}{Oxford5k}   \\
      				\hline
      				64  &       16,384 & 49.6 &       24.3\\
      				64  &   {\scriptsize PCA$\rightarrow$}\,4,096  &\textbf{52.6}     &   \textbf{25.1}   \\
      				64  &   {\scriptsize PCA$\rightarrow$}\,2,048  &{51.8}   &  24.3    \\
      				64  &  {\scriptsize PCA$\rightarrow$}\,1,024  & 51.3  &      23.4 \\
      				64  & {\scriptsize PCA$\rightarrow$}\,\,\,\,\,512 & 48.2    &  21.7     \\
      				64  &{\scriptsize PCA$\rightarrow$}\,\,\,\,\,256 & 45.9    &    20.3   \\
      				64  &  {\scriptsize PCA$\rightarrow$}\,\,\,\,\,128 & 44.6     &  19.1    \\
      				64  &   {\scriptsize PCA$\rightarrow$}\,\,\,\,\,\,\,\,64    & 42.9   &   17.2    \\
      				%64 &   \begin{footnotesize}$\xrightarrow{PCA}$\end{footnotesize}\quad\,    32  &   37.8    &   14.8    \\
      					\bottomrule[1.1pt]
      			\end{tabular}}
      		\end{table}

  In Table \ref{tab3} we also  report the performance of our derivation of the BMM-FV
  varying the number $K$ of Bernoulli mixture components %(Table\ref{tab:fvBmm}) 
  and investigating the impact of the PCA
  dimensionality reduction in the case of $K=64$. %(Table \ref{tab:PCA}).
  
  In Table (\ref{tab:fvBmm}) we can see that with the Holidays
  dataset, the mAP grows from \emph{32.0\%} when using only 4
  mixtures to \textbf{54.7\%} when using $K=512$. On Oxford5k, mAP
  varies from \emph{14.3\%} to \textbf{27.4\%}, respectively, for
  $K=4$ and $K=512$.
  
  Table (\ref{tab:PCA}) shows that the best results are achieved
  when reducing the full size BMM-FV to $4,096$ with a mAP of \textbf{52.6\%} for
  Holidays and \textbf{25.1\%} for Oxfrod5k.
  
  %table
  
  \paragraph{Analysis of the results}
            
		      Summing up, the results show that in the context of binary local features the  BMM-FV outperforms the compared aggregation methods, namely \h{the} BoW, \h{the} VLAD and \h{the} GMM-FV.
              The performance of the BMM-FV is an increasing function of the number $K$ of Benoulli mixtures. However, for large $K$, the improvement tends to be smaller and the dimensionality of the FV becomes very large (e.g. $65,536$ dimensions using $K=256$). Hence, for high values of $K$, the benefit of the improved accuracy is not worth the computational overhead (both for the BMM estimation and for the cost of storage/comparison of FVs).
              %The improvements tend to be smaller for large $K$ and, in addiction, for high value of $K$ the dimensionality of the FV becomes very large (e.g. $32,768$ dimensions for $K=128$) so for high value of $K$ the benefit of the improved accuracy is not worth the computational overhead (both for the cost of BMM estimation and FV storage/comparison).
  
              The PCA reduction of BMM-FV is effective since it can provide a very compact image signature %(even smaller than one single local feature) 
              with just a slight loss in accuracy, as shown in the case of $K=64$ (Table \ref{tab:PCA}). Dimension reduction does not necessarily reduce the accuracy. Conversely, limited reduction tend to improve the retrieval performance of the FV representations. %
  
              For the computation of VLAD, the $k$-means results are more effective than $k$-majority/$k$-medoids clustering, since the use of non-binary centroids gives more discriminant information during the computation of the residual vectors used in VLAD.
  
              For the BoW approach, $k$-means and $k$-majority performs equally better than $k$-medoids.
              However, the $k$-majority is preferable in this case because  the cost of the quantization process is significantly reduced by using the Hamming distance, rather than Euclidean one, for the comparison between centroids and binary local features.
  
              Both BMM-FV and VLAD, with only $K=64$, outperform BoW. However, as happens for non-binary features (see Table \ref{tab:res-literature}), the loss in accuracy of BoW representation is comparatively lower when the variability of the images is limited, as for the Oxford5k dataset. %\cite{jegou12}
              %Overall one can observe that BMM-FV and VLAD with only $K=64$ outperform BoW representation. However, as happens for non-binary features \cite{jegou12}, BoW is comparatively better when the variability of the images is limited, as for the Oxford5k dataset.
  
              As expected, BMM-FV outperforms  GMM-FV, since the probability distribution of binary local features is better described using mixtures of Bernoulli rather than mixtures of Gaussian.
                The results of our experiments also show that the use of BMM-FV is still effective even if compared with the direct matching strategy.
                In fact, the retrieval performance of BMM-FV on Oxford5k is just slightly worse than traditional direct matching of local feature, while on INRIA Holidays the BMM-FV even outperforms the direct matching result.
                
  For completeness, in Table \ref{tab:res-literature}, we also
    report the results of the same base-line encodings approaches applied to
    non-binary features (both full-size SIFT and PCA-reduced to 64
    components) taken from literature \cite{jegou10:VLAD,jegou12}. As
    expected, aggregation methods in general exhibit better performance in
    combination with SIFT/SIFTPCA then with ORB, expecially for the Oxford5k dataset. 
    %In the following, we then focus on results on binary features,
   However, it is worth noting that  on the INRIA Holidays the BMM-FV outperforms the BoW on SIFT/SIFTPCA and reach similar performance of the FV built upon SIFTs. 
%    In fact, SIFTPCA reaches a mAP of 43.7 on INRIA Holidays and 35.4,
%    on Oxford5k. Also in this case, we are slightly better in one case
%    and worse in the other.

The FV and VLAD get considerable benefit from performing PCA of SIFT local descriptors before the aggregation phase as the PCA rotation decorrelate the descriptors components. This suggest that techniques, such as VLAD with k-means and GMM-FV, which treat binary vectors as real-valued vectors, may also benefit from the use of PCA before the aggregation phase.
  
  In conclusion, it is
      important to point-out that there are several applications where
      binary features need to be used to improve efficiency, at the cost
      of some effectiveness reduction \cite{heinly12}. We showed that in this case the use of the encodings techniques represent a valid alternative to the direct matching. 
  %    In fact, several article
  %    \cite{rublee11} shows that ORB detects and computes about 1000
  %    features from an image in about 15.3 millisecond while SURF and
  %    SIFT need respectively 217.3 and 5228.7 millisecond for the same
  %   data and number of features. 
  %  %Thus, ORB is used in application that
  %  %  require high efficiency with small computing power, as for
  %  %  instance mobile or wearable devices, since the extraction process is
  %  %  an order of magnitude faster than SURF, and over two orders than
  %  %  SIFT.
  %    %

% \subsection{Combination of BMM-FVs and CNNs}\label{sec:expCompFVandCNN}
 
 \subsection{\h{Combination of CNNs and Aggregations of Binary Local Feature}}\label{sec:expCompFVandCNN}
  \begin{figure}[tbp]
                 \centering
                 {\includegraphics[trim=20mm 95mm 15mm 80mm, width=0.8\linewidth]{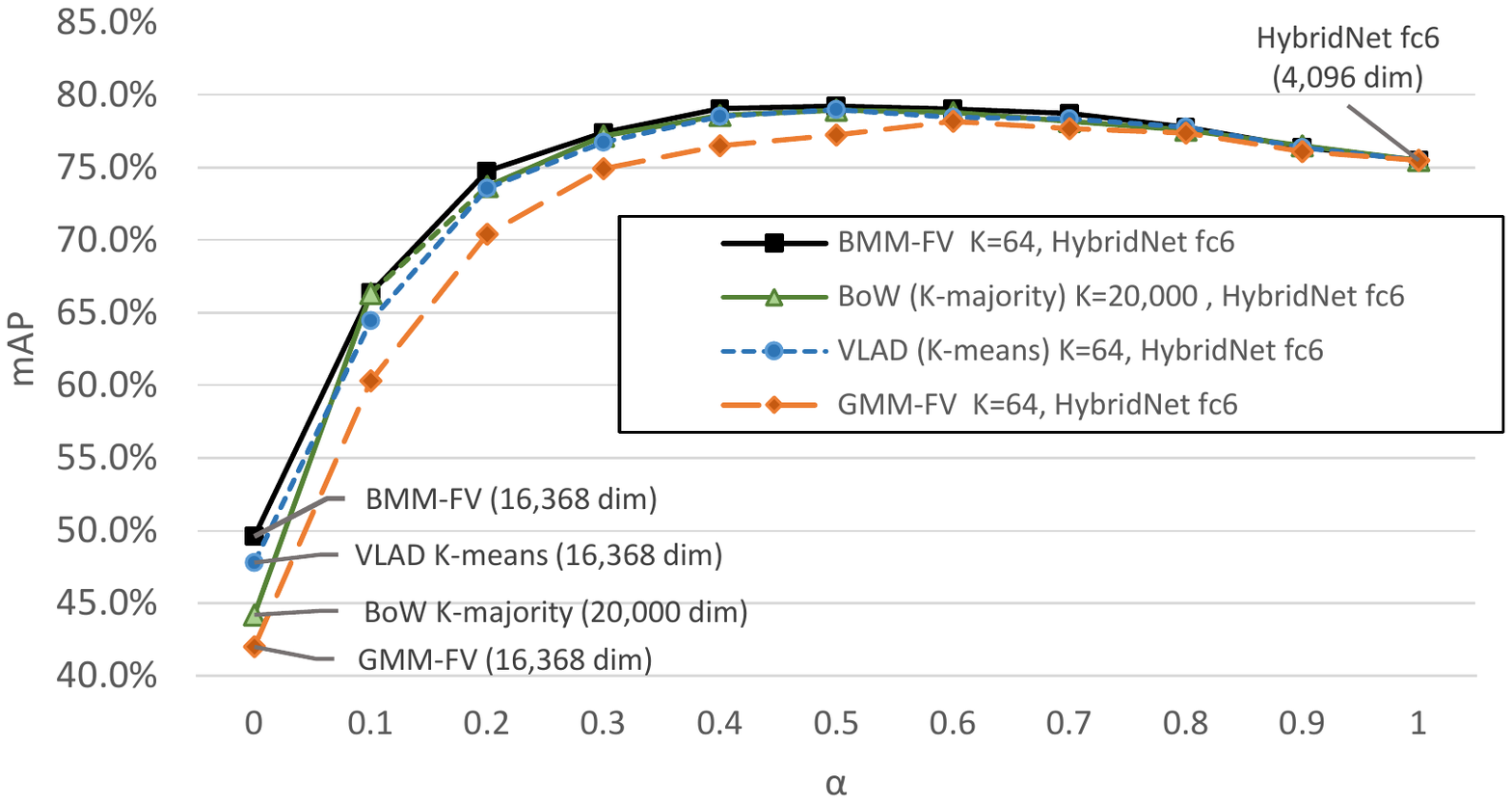}}
                 \caption{\h{Retrieval performance on the INRIA Holidays dataset of the combination of HybridNet \textit{fc6} and various aggregations of ORB binary feature (BMM-FV, VLAD, BoW, and GMM-FV). Only the full-sized descriptors are considered (i.e., no PCA) and for each aggregation technique we selected the corresponding best setting (e.g learning method) according with results reported in Table} \ref{tab:comb varying local feature}.    
                 \h{$\alpha$ is the parameter used in the combination: $\alpha = 0$ corresponds to use only the aggregated descriptor, while $\alpha=1$ correspond to use only the HybridNet feature. }
                 }
                 \label{fig:combinationCNNs-all}
                 \end{figure}

In this section we evaluate the retrieval performance of the combination of %BMM-FV and CNN features using the approach described in Section \ref{sec:cnn}. 
\h{CNN features with the aggregations of binary local feature, following the approach described in Section} \ref{sec:cnn}.
 We considered the INRIA Holidays dataset and we used the 
    the output of the first fully-connected layer (\textit{fc6}) of the HybridNet \cite{zhou2014} model as CNN feature. In fact, in \cite{chandrasekhar2015} several experiments on the INRIA Holidays have shown that \textit{HybridNet fc6} achieve better mAP result than other outputs (e.g. \textit{pool5, fc6, fc7, fc8}) of several  pre-trained CNN models: \h{the} OxfordNet \cite{simonyan2014}, \h{the} AlexNet \cite{krizhevsky2012}, \h{the} PlacesNet \cite{zhou2014} and \h{the} HybridNet itself.
 
 \h{Figure }\ref{fig:combinationCNNs-all}\h{ shows the mAP obtained by combining HybridNet \textit{fc6} with different aggregations of ORB binary features, namely the BMM-FV, the GMM-FV, the VLAD, and the BoW.
Interestingly, with the exception of the GMM-FV, the retrieval performance obtained after the combination is very similar for the various aggregation techniques. This, on the one hand confirms that the GMM-FV is not the best choice for encoding \textit{binary} features, and on the other hand, since each aggregation technique computes statistical summaries of the \textit{same set} of the local descriptors, suggests that the additional information provided by the various aggregated descriptors %during the combination 
helps almost equally to improve the retrieval performance of the CNN feature.
Thus, in the following we further investigate combinations of CNNs and the BMM-FV that, even for a shot, reaches the best performance for all the tested parameter $\alpha$.
 }
    
%             \begin{figure}[tbp]
%            \centering
%            {\includegraphics[trim=20mm 95mm 15mm 80mm, width=0.8\linewidth]{combinationFVCNNs}}
%            \caption{Graphical representation of the results reported in Table \ref{tab:comb varying local feature}, i.e. the mAP achieved with the combinations of BMM-FV and HybridNet fc6 feature for various type of binary local features. The BMM-FV representations were computed using $K=64$ mixtures of Bernoulli. $\alpha$ is the parameter used in the combination of FV and CNN.%: $\alpha = 0$ corresponds to use only FV, while $\alpha=1$ correspond to use only the HybridNet feature.
%            }
%            \label{fig:combinationFVCNNs}
%            \end{figure}
%       

     \begin{table*}[tbp]
                  \renewcommand{\arraystretch}{1}
                      \caption{Retrieval performance of various combinations of BMM-FV and HybridNet CNN feature. The BMM-FV representations were computed for three different binary local features (ORB, LATCH, and AKAZE) using $K=64$ mixtures of Bernoulli. The CNN feature was computed as the output the HybriNet fc6 layer after applying the ReLU transform. \textit{Dim} is the number of components of each vector representation. %; %please note that the dimensionality of the FV depends on the used local features. 
                      $\alpha$ is the parameter used in the combination of FV and CNN: $\alpha = 0$ corresponds to use only FV, while $\alpha=1$ correspond to use only the HybridNet feature.
                     \h{Bold numbers denote maxima in the respective column.}
                      }\label{tab:comb varying local feature}
                      \centering
           	\scriptsize %footnotesize
                  \arrayrulecolor{gray}
                      %Performance comparison of BoW, VLAD and FV aggregations of ORB binary features
                      \begin{tabular}{@{}
                  >{\centering\arraybackslash} p{0.23\textwidth} %method
                   >{\centering\arraybackslash} p{0.075\textwidth} %dim\1
                   >{\centering\arraybackslash} p{0.075\textwidth} %dim 2
                    >{\centering\arraybackslash} p{0.075\textwidth}  %dim3
                     >{\centering\arraybackslash}p{0.075\textwidth} %alpha
                    >{\centering\arraybackslash} p{0.075\textwidth}  %mAP orb
                     >{\centering\arraybackslash}p{0.075\textwidth}  %mAP latch
                     >{\centering\arraybackslash}p{0.075\textwidth}  %maP akaze
                  @{}}\toprule[1.1pt]
                  %\hline  %\toprule
                  \multirow{2}{0.23\textwidth}{\centering\textbf{Method}}	&\multicolumn{3}{c}{\textbf{Dim}} & \multirow{2}{0.07\textwidth}{\centering{$\boldsymbol{\alpha}$}}	& \multicolumn{3}{c}{\textbf{mAP}}\\
                  \cmidrule{2-4} 	\cmidrule{6-8} 
                  		&\scriptsize 	ORB	&	\scriptsize  LATCH	&	\scriptsize  AKAZE	&		&	\scriptsize 	ORB	&\scriptsize 	LATCH	&\scriptsize 	AKAZE	\\	\midrule[1.1pt]
                  BMM-FV (K=64) 	&	16,384	&	16,384	&	32,768	&	0	&		49.6 	&	46.3 	&	43.7 	\\	
                  \midrule
                  \multirow{9}{0.25\textwidth}{\centering{Combination  of \\ \textit{BMM-FV (K=64)}\\  and\\ \textit{HybridNet fc6}}}	&	\multirow{9}{0.08\textwidth}{\centering 20,480}&	\multirow{9}{0.08\textwidth}{\centering 20,480}&\multirow{9}{0.08\textwidth}{\centering 36,864}	&	0.1	&		66.4 	&	64.7 	&	59.2 	\\	
                  	&		&		&		&	0.2	&		74.8  		&	73.8  			&	68.7  	\\	
                  	&		&		&		&	0.3	&		77.4  		&	76.8  			&	74.3  	\\	
                  	&		&		&		&	0.4	&		79.1  		&	77.5  			&	77.3  	\\	
                  	&		&		&		&	0.5	&	\textbf{79.2  }	&	78.3  			&	78.0  	\\	
                  	&		&		&		&	0.6	&		79.0  		&	\textbf{78.5  }	&	\textbf{79.2  }	\\	
                  	&		&		&		&	0.7	&		78.7  		&	77.7  			&	78.7  	\\	
                  	&		&		&		&	0.8	&		77.8  		&	76.7  			&	77.5  	\\	
                  	&		&		&		&	0.9	&		76.4  		&	76.3  			&	76.2  	\\	
                  \midrule
                  HybridNet fc6	&	\multicolumn{3}{c}{4,096}							&	1	&	\multicolumn{3}{c}{75.5 }	\\	%
                  \bottomrule[1.1pt]
                  %\hline
                      \end{tabular}
                  \end{table*}

    In Table \ref{tab:comb varying local feature} %and Figure \ref{fig:combinationFVCNNs} 
    we report the mAP obtained combining the HybridNet fc6 feature with the BMM-FV computed for three different kind of binary local features, namely ORB, LATCH and AKAZE, using $K=64$ mixtures of Bernoulli.
    It is worth noting that all the three BMM-FVs %, computed on ORB, LATCH, and AKAZE, 
    give a similar improvement when combined with the HybridNet fc6, although they have rather different mAP results (see first row of Table \ref{tab:comb varying local feature}) which are substantially lower than that of CNN (last row of Table \ref{tab:comb varying local feature}). The intuition is that the additional information provided by using a specific BMM-FV rather than using the CNN feature alone, do not depend very much on the used binary feature.

     For each tested BMM-FV seems that exist an optimal $\alpha$ to be used in the convex combination (equation \eqref{eq:combination}). When ORB binary features were used, the optimal $\alpha$ was obtained around $0.5$, which correspond to give the same importance to both FV and CNN feature. For the less effective BMM-FVs built upon LATCH and AKAZE, the optimal $\alpha$ was $0.6$, which means that the CNN feature is used with slightly more importance than BMM-FV during the convex combination.
    %although the single BMM-FVs computed on ORB, LATCH, and AKAZE have rather different performance (see first row of Table \ref{tab:comb varying local feature}), they all give a similar contribute when combined with CNN feature. 
    
    The use of ORB or AKAZE led to obtain the best performance that was $\textbf{79.2\%}$ of mAP. %, using respectively  $\alpha=0.5$ and $\alpha=0.6$. 
     This results in a relative improvement of $\textbf{4.9\%}$ respect to the single use of the CNN feature, that in our case was $\textbf{75.5\%}$. So we obtain the same relative improvement of \cite{chandrasekhar2015} but using a less expensive FV representation. 
     Indeed, in \cite{chandrasekhar2015} the fusion of HybridNet \textit{fc6} and a FV computed on 64-dimensional PCA-reduced SIFTs, using $K=256$ mixtures of Gaussian, have led to obtain a relative improvement of ${4.9\%}$ respect to the use of the CNN feature alone (see also Table \ref{tab:comparison}).
     
     \h{However, the cost for integrating traditional FV built upon SIFTs with CNN features may be considered too high, especially for systems that need to process image in real time. For example, according to} 
     \cite{heinly12}
     \h{and as showed in the table} \ref{tab:efficiencyComparison}, 
     \h{the SIFTs extraction (about $2,000$ features per image), the PCA-reduction to $D=64$ dimensions, and the FV aggregation with $K=256$ requires more than $1.3$ seconds per image, while the CNN feature extraction is 4 times faster (i.e., about $300$ ms per image). 
     On the other hand, extracting ORB binary features (about $2,000$ features per image, each of dimension $D=256$) and aggregating them using a BMM-FV with $K=64$ requires less than $190$ ms that is in line with the cost of CNN extraction ($300$ ms). % Thus, the BMM-FV computation can be executed in parallel with the CNN extraction without significant increasing in the processing time.
     In our tests, the cost for integrating the already extracted BMM-FV and the CNN features was negligible in the search phase, using a sequential scan to search a dataset, also thanks to the fact that both BMM-FV and CNN features are computed using the not too costly Euclidean distance.
    % Also in the search phase, since both BMM-FV and CNN features are competed using the not too costly Euclidean distance, we haven't observed  
     }
          
    % it has been also shown that combining the  HybridNet \textit{fc6} with FV representation led to improve the mAP obtained using just the CNN representation. In particular, authors of \cite{chandrasekhar2015} used a FV computed on 64-dimensional PCA-reduced SIFTs, using $K=256$ mixture of Gaussians, which results in a $32,768$ dimensional vector. In their experiments on INRIA Holidays, the fusion of CNN and FV let to obtain a relative improvement of $\textbf{4.9\%}$ respect to the use of the CNN feature alone.
    % %In their experiments the achieved mAP on INRIA Holidays was $63\%,81\%,85\%$ respectively for the FV, HybridNet \textit{fc6}, and the fusion of latter two. Thus, the combination of CNN and FV let to obtain a relative improvement of $4.9\%$ respect to the single use of the CNN features.
    % However, that fusion uses a large number $K$ of Gaussians and requires the extraction of the costly SIFT. Since the extraction of binary local features can be up two times faster than SIFT (see \cite{rublee11,Miksik2012}), in the following we analyze how the performance of CNN feature can be improved by combining it with the less costly BMM-FV built upon binary local features.
    
           \begin{figure}[tbp]
                          \centering
                          {\includegraphics[trim=20mm 90mm 15mm 80mm, width=0.8\linewidth]{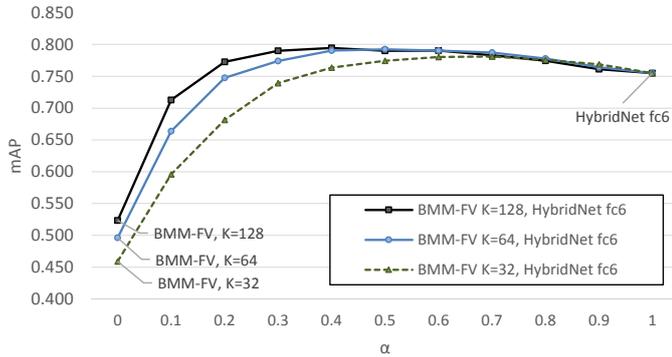}}
                          \caption{Retrieval performance of the combinations of BMM-FV and HybridNet \textit{fc6} for various number $K$ of Bernoulli mixtures. The BMM-FVs were computed using ORB binary features.
                          % The horizontal red line shows the overall maximum mAP that was obtained (i.e., $79.5\%$). 
                          $\alpha$ is the parameter used in the combination: $\alpha = 0$ corresponds to use only FV, while $\alpha=1$ correspond to use only the HybridNet feature.
                          }
                          \label{fig:bmm-fvORB}
                          \end{figure}

    Since as observed in \cite{levi15_LATCH,akaze} the ORB extraction is faster than LATCH and AKAZE, in the following we focus just on ORB binary feature.
    In figure \ref{fig:bmm-fvORB} we show the results obtained by combining HybridNet \textit{fc6}  with the BMM-FVs obtained using $K=32,64,128$. We observed that the performance of the CNN feature is improved also when it is combined with the less effective BMM-FV built using $K=32$ Bernoulli. The BMM-FV with $K=128$ achieve the best effectiveness (mAP of $\textbf{79.5\%}$) for $\alpha=0.4$. However, since the cost for computing and storing  FV increase with the number $K$ of Bernoulli, the improvement obtained using $K=128$ respect to that of $K=64$ doesn't worth the extra cost of using a bigger value of $K$. 

   The BMM-FV with $K=64$ is still high dimensional, so to reduce the cost of storing and comparing FV, we also evaluated the combination after the PCA dimensionality reduction.
           As already observed, limited dimensionality reduction tends to improve the accuracy of the single FV representation. In fact, the BMM-FV with $K=64$ achieved a mAP of $52.6\%$ when reduced from $16,384$ to $4,096$ dimensions. However, as shown in Table \ref{tab:pcanopca} and Table \ref{tab:comparison}, when the PCA-reduced version of the BMM-FV was combined with HybriNet \textit{fc6}, the overall relative improvement in mAP was $3.9\%$, which is less than that obtained using the full-sized BMM-FV. These result is not surprising given that after the dimensionality reduction we may have a loss of the additional information provided by the FV representation during the combination with the CNN feature. 
   
    Finally, in Table  \ref{tab:comparison} we summarizes the relative improvement achieved by combining BMM-FV and HybriNet \textit{fc6}, and we compare the obtained results with the relative improvement achieved in \cite{chandrasekhar2015}, where the more expensive FV built upon SIFTs was used. We observed that BMM-FV led to achieve similar or even better relative improvements with an evident advantage from the computational point of view, because it uses binary local features.% and smaller number $K$ of mixtures.
 	
 	 \begin{table*}[tbp]
 	             \renewcommand{\arraystretch}{1}
 	                 \caption{Comparison of the results obtained combining HybridNet fc6 feature with the full-sized and the PCA-reduced version of the BMM-FV. The BMM-FV was computed on ORB binary feature using $K=64$ mixtures of Bernoulli. \textit{Dim} is the number of components of each vector representation. $\alpha$ is the parameter used in the combination of FV and CNN.
 	                \h{Bold numbers denote maxima in the respective column.}
 	                 }\label{tab:pcanopca}
 	                 \centering
 	              	\scriptsize %footnotesize
 	             	\arrayrulecolor{gray}
 	             	    %Performance comparison of BoW, VLAD and FV aggregations of ORB binary features
 	             	    \begin{tabular}{@{}
 	             	>{\centering\arraybackslash} p{0.23\textwidth} %method															
 	             	 >{\centering\arraybackslash} p{0.1\textwidth} %FV full dim															
 	             	 >{\centering\arraybackslash} p{0.1\textwidth} %FV PCA-reduced to 4096															
 	             	   >{\centering\arraybackslash}p{0.1\textwidth} %alpha															
 	             	  >{\centering\arraybackslash} p{0.1\textwidth}  %mAP 1															
 	             	   >{\centering\arraybackslash}p{0.1\textwidth}  %mAP2															
 	             	@{}}\toprule[1.1pt]															
 	             	%\hline  %\toprule															
 	             	\multirow{2}{0.23\textwidth}{\centering\textbf{Method}}	&	\multicolumn{2}{c}{\textbf{Dim}} 			&	\multirow{2}{0.1\textwidth}{\centering{$\boldsymbol{\alpha}$}}		 	& \multicolumn{2}{c}{\textbf{mAP}}\\				\cmidrule{2-3} 	\cmidrule{5-6} 	
 	             																
 	             		&	\scriptsize{FV full dim}	& \scriptsize{FV PCA-reduced}	&		&	\scriptsize 	 FV full dim	&\scriptsize 	FV PCA-reduced			\\	\midrule[1.1pt]
 	             	BMM-FV (K=64) 	&	16,384	&	4,096	&	0	&		49.6 	&	52.6	\\	
 	             	\midrule											
 	             	\multirow{9}{0.25\textwidth}{\centering{Combination  of \\ \textit{BMM-FV (K=64)}\\  and\\ \textit{HybridNet fc6}}}	&	\multirow{9}{0.085\textwidth}{\centering 20,480}	&	\multirow{9}{0.085\textwidth}{\centering 8,192}	&	0.1	&		66.4  		&	66.3	 	\\
 	             		&		&		&	0.2	&		74.8  		&	73.9	 	\\
 	             		&		&		&	0.3	&		77.4  		&	77.3	 	\\
 	             		&		&		&	0.4	&		79.1  		&	\textbf{78.5}	 	\\
 	             		&		&		&	0.5	&		\textbf{79.2  }		&	78.4	 	\\
 	             		&		&		&	0.6	&		79.0  		&	{78.5	 }	\\
 	             		&		&		&	0.7	&		78.7  		&	78.1	 	\\
 	             		&		&		&	0.8	&		77.8  		&	77.7	 	\\
 	             		&		&		&	0.9	&		76.4  		&	76.4	 	\\
 	             	\midrule																													
 	             	HybridNet fc6	&	\multicolumn{2}{c}{4,096}					&	1	&	\multicolumn{2}{c}{75.5  }	\\	%			
 	             	\bottomrule[1.1pt]
 	             	%\hline
 	                 \end{tabular}
 	             \end{table*}
 	             
 	               \begin{table}[tbp]
 	                	\centering
 	              	\scriptsize %footnotesize
 	                			\caption{Relative mAP improvement obtained after combining FV with HybridNet \textit{fc6}. Each relative improvements was computed respect to the use of the CNN feature alone, that is: $\left(\text{mAP}_\text{after combination}-\text{mAP}_\text{HybridNet fc6}\right)/\text{mAP}_\text{HybridNet fc6}$. The relative improvements obtained using the FV computed on 64-dimensional PCA-reduced SIFTs (SIFTPCA64) was computed according to the results reported in \cite{chandrasekhar2015}. }\label{tab:comparison}
 	                	{%
 	                				\begin{tabular}{
 	                						>{\raggedright\arraybackslash}p{0.2\columnwidth}
 	                						>{\centering\arraybackslash}p{0.17\columnwidth}
 	                						>{\centering\arraybackslash}p{0.1\columnwidth}
 	                						>{\centering\arraybackslash}p{0.15\columnwidth} 
 	                						>{\centering\arraybackslash}p{0.17\columnwidth} 
 	                					@{}}\toprule[1.1pt]
 	                	\multirow{2}{0.15\textwidth}{\centering\textbf{FV\\ method}}	&	
 	                	\multirow{2}{0.2\textwidth}{\centering\textbf{Local \\Feature}}	&
 	                	\multirow{2}{0.1\textwidth}{\centering\textbf{K}}	&	
 	                	\multirow{2}{0.15\textwidth}{\centering\textbf{dim}}		&	
 	                	\multirow{2}{0.17\textwidth}{\centering\textbf{Relative improvement}}	\\
 	                	\\\midrule[1.1pt]
 	                	BMM-FV%$_{\mu}$
 	                		&	ORB	&	128	&	32,768		&	5.2	 	\\	
 	                	BMM-FV%$_{\mu}$
 	                		&	ORB	&	64	&	16,384		&	4.9	 	\\	
 	                	BMM-FV%$_{\mu}$
 	                		&	AKAZE	&	64	&	32,768		&	4.9	 	\\	
 	                	BMM-FV%$_{\mu}$
 	                		&	LATCH	&	64	&	16,384		&	4.0	 	\\
 	                	BMM-FV%$_{\mu}$ 
 	                	+ PCA	&	ORB	&	64	&	4,096	&	3.9	 	\\	
 	                	BMM-FV%$_{\mu}$
 	                		&	ORB	&	32	&	8,192		&	3.5	 	\\	\midrule
 	                	 FV%$_{\mu\sigma}$ 
 	                	\cite{chandrasekhar2015}	&	SIFTPCA64	&	256	&	32,768		&	4.9	  	\\
 	                	
 	                		\bottomrule[1.1pt]
 	                				\end{tabular}}		
 	                		\end{table}
 	                		
 	 \subsection{\h{Large-Scale Experiments}}\label{sec:largescale}
 	       
 	   		   \begin{table*}[tbp]%[tbp]
 	   		      \renewcommand{\arraystretch}{1}
 	   		          \caption{
 	   		          \h{Comparison of the results obtained combining HybridNet fc6 feature and BMM-FV on the INRIA Holidays dataset with the distractor dataset MIRFlickr-1M. 
 	   				The results related to the INRIA Holidays alone are reported from Table}  \ref{tab:pcanopca} \h{for reference. 
 	   				The BMM-FV was computed on ORB binary feature using $K=64$ mixtures of Bernoulli; both full-sized and the PCA-reduced features are considered. 
 	   				\textit{Dim} is the number of components of each vector representation. $\alpha$ is the parameter used in the combination of FV and CNN.
 	   				Bold numbers denote maxima in the respective column. The last row reports the maximum relative mAP improvement obtained after combining FV with HybridNet \textit{fc6}; relative improvements are computed respect to the use of the CNN feature alone, that is $\left(\text{mAP}_\text{after combination}-\text{mAP}_\text{HybridNet fc6}\right)/\text{mAP}_\text{HybridNet fc6}$.  }
 	   		          }\label{tab:largescaleALL}
 	   		          \centering
 	   		         \scriptsize %footnotesize
 	   		 %	\arrayrulecolor{gray}
 	   		      	  
 	   		      	    \begin{tabular}{%@{}			
 	   		      	>{\centering\arraybackslash} p{0.18\textwidth}| %method															
 	   		      	 >{\centering\arraybackslash} p{0.08\textwidth} %FV full dim															
 	   		      	 >{\centering\arraybackslash} p{0.08\textwidth} %FV PCA-reduced to 4096															
 	   		      	   >{\centering\arraybackslash}p{0.06\textwidth} %alpha															
 	   		      	  >{\centering\arraybackslash} p{0.065\textwidth}  %mAP1 1															
 	   		      	   >{\centering\arraybackslash}p{0.065\textwidth}  %mAP2	
 	   		      	   	   >{\centering\arraybackslash}p{0.0051\textwidth}  %spazio
 	   		      	  >{\centering\arraybackslash}p{0.065\textwidth}  %mAP3	
 	   		      	   >{\centering\arraybackslash}p{0.065\textwidth}%mAP4													
 	   		      %	@{}
 	   		      	}\toprule[1.1pt]										      											
 	   		      		\multirow{3}{0.18\textwidth}{\centering\textbf{Method}}	& \multicolumn{2}{c}{\textbf{Dim}}&	\multirow{3}{0.06\textwidth}{\centering{$\boldsymbol{\alpha}$}}	&	\multicolumn{5}{c}{\textbf{mAP}}\\ 	\cmidrule{2-3}	\cmidrule{5-9}       	
 	   		      		&	{\multirow{2}{0.08\textwidth}{\centering\scriptsize{FV full dim}}}	& 	{\multirow{2}{0.08\textwidth}{\scriptsize{FV PCA-reduced}}}	&		&	\multicolumn{2}{c}{\scriptsize{FV full dim}}	& &	\multicolumn{2}{c}{\scriptsize{FV PCA-reduced}}			\\	\cmidrule{5-6}	\cmidrule{8-9} 	
 	   		      		&		&		&		&	\rot{{Holidays}}	&	\rot{\multirow{1}{0.06\textwidth}{Holidays+ MIRFlickr}}	&&	 \rot{{Holidays}}	&	\rot{\multirow{1}{0.06\textwidth}{Holidays+ MIRFlickr}}	\\      	
 	   		      		\midrule	
 	   		      	BMM-FV (K=64) 	&	16,384	&	16,384	&	0	&	49.6	&	31.0	&&		52.6	&	34.9	\\	\midrule	
 	   		      	\multirow{9}{0.18\textwidth}{\centering{Combination  of \\ \textit{BMM-FV (K=64)}\\  and\\ \textit{HybridNet fc6}}}	&	20,480	&	8,192	&	0.1	&	66.4	&	47.0	&&		66.3	&	50.7	\\		
 	   		      		&		&		&	0.2	&	74.8	&	59.3	&&		73.9	&	61.9	\\		
 	   		      		&		&		&	0.3	&	77.4	&	64.0	&&	77.3	&	65.6	\\		
 	   		      		&		&		&	0.4	&	79.1	&	\textbf{67.1}	&&		\textbf{78.5}	&	\textbf{67.2}	\\		
 	   		      		&		&		&	0.5	&\textbf{79.2}&	66.5	&&		78.4	&	66.9	\\		
 	   		      		&		&		&	0.6	&	79.0	&	65.7	&&		78.5	&	65.7	\\		
 	   		      		&		&		&	0.7	&	78.7	&	64.4	&&		78.1	&	64.4	\\		
 	   		      		&		&		&	0.8	&	77.8	&	62.5	&&		77.7	&	62.8	\\		
 	   		      		&		&		&	0.9	&	76.4	&	60.7	&&		76.4	&	60.8	\\	\midrule	
 	   		      	HybridNet fc6 	&	\multicolumn{2}{c}{4,096}	&	1	&	75.5	&	59.1	&&		75.5	&	59.1	\\	\bottomrule[1.1pt]	\\
 	   			\multicolumn{4}{r}{	\textbf{Maximum relative mAP improvement} $\rightarrow$}& 4.9\%&13.4\%&&4.0\%&13.7\%
 	   		      	
 	   		          \end{tabular}
 	   		          
 	   		      \end{table*}
 	   		   
   		 \h{ In order to evaluate the behavior of feature combinations on a large scale, we have used a set of up to one million images. More precisely, as in }\cite{jegou08}\h{, we merged the INRIA Holidays dataset with a public large-scale dataset (MIRFlickr-1M} \cite{huiskes08}\h{) used as distraction set; the mAP was measured using the Holidays ground-truth.}
   		  
   		  \h{Table} \ref{tab:largescaleALL}\h{ reports results obtained using both the BMM-FV alone and the combinations with the \textit{HybridNet fc6} CNN feature.  Given the results reported in the previous section we focus on the BMM-FV encoding of ORB binary features. All the feature combinations show an improvement with respect to the single use of the CNN feature (mAP of $59.1\%$) or BMM-FV (mAP of $31.0\%/34.9\%$ respectively using the full length/PCA-reduced descriptor). 	  
   		  This reflects the very good behavior of feature combinations also in the large-scale case.}   		  
   		
   		  \h{The mAP reaches a maximum using $\alpha$ between 0.4 and 0.5, that is giving (quite) the same weight to BMM-FV and CNN feature during the combination. The results obtained using the full length BMM-FV and the PCA-reduced version are similar. The latter performs slightly  better and achieved a maximum of $\textbf{67.2\%}$ of mAP that correspond to $\textbf{13.7\%}$ of relative mAP improvement respect to use the CNN feature alone. 
	   	  It is worth noting that the relative mAP improvement obtained in the large-scale setting is much greater than that obtained without the distraction set. This suggests that the information provided by the BMM-FV during the combination helps in discerning the visual content of images particularly in presence of distractor images.
	   	  
		   Since the computational time of extracting binary features is much faster than others, the computational gain of combining CNN features with BMM-FV encodings of ORB over traditional FV encodings of SIFT is 
		   especially notable in the large-scale scenario. For example, the process for extracting SIFTs from the INRIA Holidays+ MIRFlickr dataset ($1,001,491$ images) would have required more than 13 days (about 1,200 ms per image) while ORB extraction took less than 8 hours (about 26 ms per image). 
		   }
   		% \todo{se fai in tempo inserisci immagini qualitative}
%   		 For large-scale retrieval experiments we used
%   		  combined with the 1 million MIR-FLICKR distractor data se
   		 
%   		 Results for various queries are presented in Fig.
%   		 16
%   		 .One
%   		 can observe that the scenes returned are taken from very dif-
%   		 ferent viewpoints and orientations. The last three rows show
%   		 that some images from the
%   		 Flickr1M
%   		 dataset (marked as FFP,
%   		 false
%   		 false positive) are actually relevant to the query image.
%   		In this section we conduct experiments on large scale
%   		They are counted as false positives and artificially decrease
%   		the results in terms of mAP given in Fig.
%   		15
%   		. Figure
%   		17
%   		shows
%   		for examples for which HE and WGC improve the quality of
%   		the ranking significantly

\section{Conclusion}
\label{sec:conclusion}
  %%PAST SIMPLE: what you did in your RESEARCH.
  %%PRESENT PERFECT:to describe what you have done in the PAPER itself [during the writing process]] (di solito con verbi tipo describe,outline, present, propose, show,highlight.)
  %% WILL: to refer to future research
  %%      will is only used to refer to a real future. There is no need to use will to refer to what you will do later in the paper.
  %%  Do NOT use will to refer to the implications of your results (e.g. The present findings will ..)-> usa should / may
  
  Motivated by recent results obtained on one hand with the use of
  aggregation methods applied to local descriptors\h{,} and on the
  other with the definition of binary local features, this paper
  has performed an extensive comparisons of techniques that mix the two approaches by using aggregation methods on binary
  local features. 
  The use of aggregation methods on binary local features is
  %justified 
  \h{motivated}
  by the need for increasing efficiency and reducing computing resources for image matching on a large scale, at the
  expense of some degradation in the accuracy of retrieval
  algorithms. Combining the two approaches lead
 to execute image retrieval on a very large scale and reduce the
  cost for feature extraction and representation.
%   Our experiments show that whenever binary local feature are used
%   to represent images then it is convenient to aggregate them. In
%   facts, effectiveness of the aggregations is in line with that
%   obtained using direct matching of binary feature. At the same time,
%   aggregation techniques, jointly with the use of approximate and
%   similarity search techniques \cite{ZADB06Similarity}, offer much
%   higher scalability.
  Thus we expect that the results of our empirical evaluation are useful for people working with binary local descriptors.
 
 Moreover, we investigated how 
 \h{aggregations of binary local features work}
 %the Fisher Vector obtained by aggregating binary local features (BMM-FV)  works
 in conjunction with the CNN pipeline in order to improve the latter retrieval performance. We showed that the BMM-FV built upon ORB binary features can be profitable use to this scope, even if a relative small number of Bernoulli is used. In fact, the relative improvement in the retrieval performance obtained \h{combining CNN features with} %using 
 the BMM-FV is similar to that previously obtained in \cite{chandrasekhar2015} \h{where a combination of the CNN features with} %using
 the more expensive FV built on SIFT \h{was proposed}.
 \h{Experimental evaluation on large scale confirms the effectiveness and scalability of our proposal.}
 
  It is also worth mentioning that the BMM-FV approach is
  very general and could be applied to any binary feature.
  Recent works based on CNNs suggest that binary features aggregation technique could be further
  applied to deep features. In fact, on one hand, local
  features based on CNNs, aggregated with VLAD and FV approaches,
  have been proposed to obtain robustness to geometric deformations \cite{Ng_2015_CVPR_Workshops,Uricchio_2015_ICCV_Workshops}. On
  the other hand, binarization of global CNN features have been also
  proposed in \cite{Lin_2015_CVPR_Workshops,Lai_2015_CVPR}. Thus, as
  a future work, we plan to test the  BMM-FV approach over binary
  deep local descriptors leveraging on the local and binary
  approaches mentioned above.

  \begin{acknowledgements}
This work was partially founded by: EAGLE, Europeana network of Ancient Greek and Latin Epigraphy, co-founded by the European Commission, CIP-ICT-PSP.2012.2.1 - Europeana and creativity, Grant Agreement n. 325122; and Smart News, Social sensing for breakingnews, co-founded by the Tuscany region under the FAR-FAS 2014 program, CUP CIPE D58C15000270008.
  \end{acknowledgements}

\bibliographystyle{spmpsci}      % mathematics and physical sciences
\bibliography{bib,bibDaUniformare}

\begin{thebibliography}{10}
\providecommand{\url}[1]{{#1}}
\providecommand{\urlprefix}{URL }
\expandafter\ifx\csname urlstyle\endcsname\relax
  \providecommand{\doi}[1]{DOI~\discretionary{}{}{}#1}\else
  \providecommand{\doi}{DOI~\discretionary{}{}{}\begingroup
  \urlstyle{rm}\Url}\fi

\bibitem{akaze}
Alcantarilla, P.F., Nuevo, J., Bartoli, A.: Fast explicit diffusion for
  accelerated features in nonlinear scale spaces.
\newblock In: In British Machine Vision Conference (BMVC) (2013)

\bibitem{Amato2016}
Amato, G., Falchi, F., Gennaro, C., Vadicamo, L.: Deep Permutations: Deep
  Convolutional Neural Networks and Permutation-Based Indexing, pp. 93--106.
\newblock Springer International Publishing, Cham (2016).
\newblock \doi{10.1007/978-3-319-46759-7_7}.
\newblock \urlprefix\url{http://dx.doi.org/10.1007/978-3-319-46759-7_7}

\bibitem{amato16}
Amato, G., Falchi, F., Vadicamo, L.: How effective are aggregation methods on
  binary features?
\newblock In: Proceedings of the 11th Joint Conference on Computer Vision,
  Imaging and Computer Graphics Theory and Applications, vol.~4, pp. 566--573
  (2016)

\bibitem{amato16:JOCCH}
Amato, G., Falchi, F., Vadicamo, L.: Visual recognition of ancient inscriptions
  using convolutional neural network and fisher vector.
\newblock J. Comput. Cult. Herit. \textbf{9}(4), 21:1--21:24 (2016).
\newblock \doi{10.1145/2964911}.
\newblock \urlprefix\url{http://doi.acm.org/10.1145/2964911}

\bibitem{arandjelovic12:rootsift}
Arandjelovic, R., Zisserman, A.: Three things everyone should know to improve
  object retrieval.
\newblock In: Computer Vision and Pattern Recognition (CVPR), 2012 IEEE
  Conference on, pp. 2911--2918 (2012)

\bibitem{arandjelovic13:allAbVALD}
Arandjelovic, R., Zisserman, A.: All about {VLAD}.
\newblock In: Computer Vision and Pattern Recognition (CVPR), 2013 IEEE
  Conference on, pp. 1578--1585 (2013).
\newblock \doi{10.1109/CVPR.2013.207}

\bibitem{babenko2014neural}
Babenko, A., Slesarev, A., Chigorin, A., Lempitsky, V.: Neural codes for image
  retrieval.
\newblock In: Computer Vision--ECCV 2014, pp. 584--599. Springer (2014).
\newblock \doi{10.1007/978-3-319-10590-1_38}.
\newblock \urlprefix\url{http://dx.doi.org/10.1007/978-3-319-10590-1_38}

\bibitem{bay06}
Bay, H., Tuytelaars, T., Van~Gool, L.: Surf: Speeded up robust features.
\newblock In: A.~Leonardis, H.~Bischof, A.~Pinz (eds.) Computer Vision - ECCV
  2006, \emph{Lecture Notes in Computer Science}, vol. 3951, pp. 404--417.
  Springer Berlin Heidelberg (2006).
\newblock \doi{10.1007/11744023_32}.
\newblock \urlprefix\url{http://dx.doi.org/10.1007/11744023_32}

\bibitem{Bing-Images}
Bing images.
\newblock \urlprefix\url{http://www.bing.com/images/}

\bibitem{bishop06}
Bishop, C.M.: Pattern Recognition and Machine Learning.
\newblock Information Science and Statistics. Springer (2006)

\bibitem{boureau10}
Boureau, Y.L., Bach, F., LeCun, Y., Ponce, J.: Learning mid-level features for
  recognition.
\newblock In: Computer Vision and Pattern Recognition (CVPR), 2010 IEEE
  Conference on, pp. 2559--2566 (2010)

\bibitem{calonder10}
Calonder, M., Lepetit, V., Strecha, C., Fua, P.: Brief: Binary robust
  independent elementary features.
\newblock In: K.~Daniilidis, P.~Maragos, N.~Paragios (eds.) Computer Vision -
  ECCV 2010, \emph{Lecture Notes in Computer Science}, vol. 6314, pp. 778--792.
  Springer Berlin Heidelberg (2010)

\bibitem{chandrasekhar2015}
Chandrasekhar, V., Lin, J., Mor{\`{e}}re, O., Goh, H., Veillard, A.: A
  practical guide to cnns and fisher vectors for image instance retrieval.
\newblock CoRR \textbf{abs/1508.02496} (2015).
\newblock \urlprefix\url{http://arxiv.org/abs/1508.02496}

\bibitem{chen11}
Chen, D., Tsai, S., Chandrasekhar, V., Takacs, G., Chen, H., Vedantham, R.,
  Grzeszczuk, R., Girod, B.: Residual enhanced visual vectors for on-device
  image matching.
\newblock In: Signals, Systems and Computers (ASILOMAR), 2011 Conference Record
  of the Forty Fifth Asilomar Conference on, pp. 850--854 (2011).
\newblock \doi{10.1016/j.sigpro.2012.06.005}.
\newblock \urlprefix\url{http://dx.doi.org/10.1016/j.sigpro.2012.06.005}

\bibitem{chum07}
Chum, O., Philbin, J., Sivic, J., Isard, M., Zisserman, A.: Total recall:
  Automatic query expansion with a generative feature model for object
  retrieval.
\newblock In: Computer Vision, 2007. ICCV 2007. IEEE 11th International
  Conference on, pp. 1--8 (2007)

\bibitem{csurka04}
Csurka, G., Dance, C., Fan, L., Willamowski, J., Bray, C.: Visual
  categorization with bags of keypoints.
\newblock Workshop on statistical learning in computer vision, ECCV
  \textbf{1}(1-22), 1--2 (2004)

\bibitem{datta05}
Datta, R., Li, J., Wang, J.Z.: Content-based image retrieval: Approaches and
  trends of the new age.
\newblock In: Proceedings of the 7th ACM SIGMM International Workshop on
  Multimedia Information Retrieval, MIR '05, pp. 253--262. ACM, New York, NY,
  USA (2005)

\bibitem{delhumeau13}
Delhumeau, J., Gosselin, P.H., J{\'e}gou, H., P{\'e}rez, P.: Revisiting the
  {VLAD} image representation.
\newblock In: Proceedings of the 21st ACM International Conference on
  Multimedia, MM 2013, pp. 653--656. ACM, New York, NY, USA (2013).
\newblock \doi{10.1145/2502081.2502171}.
\newblock \urlprefix\url{http://doi.acm.org/10.1145/2502081.2502171}

\bibitem{deng2009}
Deng, J., Dong, W., Socher, R., Li, L., Li, K., Fei-Fei, L.: Imagenet: A
  large-scale hierarchical image database.
\newblock In: Computer Vision and Pattern Recognition, 2009. CVPR 2009. IEEE
  Conference on, pp. 248--255 (2009).
\newblock \doi{10.1109/CVPR.2009.5206848}

\bibitem{DeCaf}
Donahue, J., Jia, Y., Vinyals, O., Hoffman, J., Zhang, N., Tzeng, E., Darrell,
  T.: Decaf: {A} deep convolutional activation feature for generic visual
  recognition.
\newblock CoRR \textbf{abs/1310.1531} (2013).
\newblock \urlprefix\url{http://arxiv.org/abs/1310.1531}

\bibitem{galvez11}
Galvez-Lopez, D., Tardos, J.: Real-time loop detection with bags of binary
  words.
\newblock In: Intelligent Robots and Systems (IROS), 2011 IEEE/RSJ
  International Conference on, pp. 51--58 (2011)

\bibitem{vanGemert08}
van Gemert, J.C., Geusebroek, J.M., Veenman, C.J., Smeulders, A.W.: Kernel
  codebooks for scene categorization.
\newblock In: D.~Forsyth, P.~Torr, A.~Zisserman (eds.) Computer Vision - ECCV
  2008, \emph{Lecture Notes in Computer Science}, vol. 5304, pp. 696--709.
  Springer Berlin Heidelberg (2008)

\bibitem{DeepLearning2016}
Goodfellow, I., Bengio, Y., Courville, A.: Deep learning (2016).
\newblock \urlprefix\url{http://www.deeplearningbook.org}.
\newblock Book in preparation for MIT Press

\bibitem{Google-Goggles}
Google googles.
\newblock \urlprefix\url{http://www.google.com/mobile/goggles/}

\bibitem{Google-Images}
Google images.
\newblock \urlprefix\url{https://images.google.com/}

\bibitem{grana13}
Grana, C., Borghesani, D., Manfredi, M., Cucchiara, R.: {A fast approach for
  integrating ORB descriptors in the bag of words model}.
\newblock In: C.G.M. Snoek, L.S. Kennedy, R.~Creutzburg, D.~Akopian,
  D.~W\"{u}ller, K.J. Matherson, T.G. Georgiev, A.~Lumsdaine (eds.) IS\&T/SPIE
  Electronic Imaging. International Society for Optics and Photonics (2013)

\bibitem{gray98}
Gray, R.M., Neuhoff, D.L.: Quantization.
\newblock Information Theory, IEEE Transactions on \textbf{44}(6), 2325--2383
  (1998).
\newblock \doi{10.1109/18.720541}.
\newblock \urlprefix\url{http://dx.doi.org/10.1109/18.720541}

\bibitem{Hamming:1950:EDE}
Hamming, R.W.: Error detecting and error correcting codes.
\newblock The Bell System Technical Journal \textbf{29}(2), 147--160 (1950).
\newblock \doi{10.1002/j.1538-7305.1950.tb00463.x}

\bibitem{heinly12}
Heinly, J., Dunn, E., Frahm, J.M.: Comparative evaluation of binary features.
\newblock In: Computer Vision - ECCV 2012, Lecture Notes in Computer Science,
  pp. 759--773. Springer Berlin Heidelberg (2012)

\bibitem{householder64}
Householder, A.: The Theory of Matrices in Numerical Analysis.
\newblock A Blaisdell book in pure and applied sciences: introduction to higher
  mathematics. Blaisdell Publishing Company (1964)

\bibitem{huiskes08}
Huiskes, M.J., Lew, M.S.: The mir flickr retrieval evaluation.
\newblock In: MIR '08: Proceedings of the 2008 ACM International Conference on
  Multimedia Information Retrieval. ACM, New York, NY, USA (2008)

\bibitem{jaakkola98}
Jaakkola, T., Haussler, D.: Exploiting generative models in discriminative
  classifiers.
\newblock In: In Advances in Neural Information Processing Systems 11, pp.
  487--493. MIT Press (1998).
\newblock \urlprefix\url{http://dl.acm.org/citation.cfm?id=340534.340715}

\bibitem{jegou08}
J{\'e}gou, H., Douze, M., Schmid, C.: Hamming embedding and weak geometric
  consistency for large scale image search.
\newblock In: D.~Forsyth, P.~Torr, A.~Zisserman (eds.) European Conference on
  Computer Vision, \emph{LNCS}, vol.~I, pp. 304--317. Springer (2008)

\bibitem{jegou10:improvingBoW}
J{\'e}gou, H., Douze, M., Schmid, C.: Improving bag-of-features for large scale
  image search.
\newblock International Journal of Computer Vision \textbf{87}(3), 316--336
  (2010).
\newblock \doi{10.1007/s11263-009-0285-2}.
\newblock \urlprefix\url{http://dx.doi.org/10.1007/s11263-009-0285-2}

\bibitem{jegou11:PQ}
J{\'e}gou, H., Douze, M., Schmid, C.: Product quantization for nearest neighbor
  search.
\newblock Pattern Analysis and Machine Intelligence, IEEE Transactions on
  \textbf{33}(1), 117--128 (2011).
\newblock \doi{10.1109/TPAMI.2010.57}

\bibitem{jegou10:VLAD}
J{\'e}gou, H., Douze, M., Schmid, C., P{\'e}rez, P.: Aggregating local
  descriptors into a compact image representation.
\newblock In: IEEE Conference on Computer Vision \& Pattern Recognition (2010).
\newblock \doi{10.1109/CVPR.2010.5540039}

\bibitem{jegou12}
J{\'e}gou, H., Perronnin, F., Douze, M., S{\`a}nchez, J., P{\'e}rez, P.,
  Schmid, C.: Aggregating local image descriptors into compact codes.
\newblock IEEE Transactions on Pattern Analysis and Machine Intelligence
  \textbf{34}(9), 1704--1716 (2012).
\newblock \doi{10.1109/TPAMI.2011.235}

\bibitem{caffe2014}
Jia, Y., Shelhamer, E., Donahue, J., Karayev, S., Long, J., Girshick, R.,
  Guadarrama, S., Darrell, T.: Caffe: Convolutional architecture for fast
  feature embedding.
\newblock In: Proceedings of the ACM International Conference on Multimedia,
  pp. 675--678. ACM (2014).
\newblock \doi{10.1145/2647868.2654889}.
\newblock \urlprefix\url{http://doi.acm.org/10.1145/2647868.2654889}

\bibitem{kaufman87}
Kaufman, L., Rousseeuw, P.: Clustering by means of medoids.
\newblock In: Y.~Dodge (ed.) An introduction to L1-norm based statistical data
  analysis, \emph{Computational Statistics \& Data Analysis}, vol.~5 (1987)

\bibitem{krapac11}
Krapac, J., Verbeek, J., Jurie, F.: {Modeling Spatial Layout with Fisher
  Vectors for Image Categorization}.
\newblock In: {ICCV 2011 - International Conference on Computer Vision}, pp.
  1487--1494. {IEEE}, Barcelona, Spain (2011)

\bibitem{krizhevsky2012}
Krizhevsky, A., Sutskever, I., Hinton, G.E.: Imagenet classification with deep
  convolutional neural networks.
\newblock In: F.~Pereira, C.~Burges, L.~Bottou, K.~Weinberger (eds.) Advances
  in Neural Information Processing Systems 25, pp. 1097--1105. Curran
  Associates, Inc. (2012)

\bibitem{Lai_2015_CVPR}
Lai, H., Pan, Y., Liu, Y., Yan, S.: Simultaneous feature learning and hash
  coding with deep neural networks.
\newblock In: The IEEE Conference on Computer Vision and Pattern Recognition
  (CVPR) (2015)

\bibitem{lazebnik06}
Lazebnik, S., Schmid, C., Ponce, J.: Beyond bags of features: Spatial pyramid
  matching for recognizing natural scene categories.
\newblock In: Computer Vision and Pattern Recognition, 2006 IEEE Computer
  Society Conference on, vol.~2 (2006)

\bibitem{LeCun2015}
LeCun, Y., Bengio, Y., Hinton, G.: {Deep learning}.
\newblock Nature \textbf{521}(7553), 436--444 (2015).
\newblock \doi{10.1038/nature14539}

\bibitem{lee15}
Lee, S., Choi, S., Yang, H.: Bag-of-binary-features for fast image
  representation.
\newblock Electronics Letters \textbf{51}(7), 555--557 (2015)

\bibitem{leutenegger11}
Leutenegger, S., Chli, M., Siegwart, R.: Brisk: Binary robust invariant
  scalable keypoints.
\newblock In: Computer Vision (ICCV), 2011 IEEE International Conference on,
  pp. 2548--2555 (2011)

\bibitem{levi15_LATCH}
Levi, G., Hassner, T.: {LATCH:} learned arrangements of three patch codes.
\newblock CoRR \textbf{abs/1501.03719} (2015)

\bibitem{Lin_2015_CVPR_Workshops}
Lin, K., Yang, H.F., Hsiao, J.H., Chen, C.S.: Deep learning of binary hash
  codes for fast image retrieval.
\newblock In: The IEEE Conference on Computer Vision and Pattern Recognition
  (CVPR) Workshops (2015)

\bibitem{kmeans}
Lloyd, S.: Least squares quantization in pcm.
\newblock Information Theory, IEEE Transactions on \textbf{28}(2), 129--137
  (1982).
\newblock \doi{10.1109/TIT.1982.1056489}.
\newblock \urlprefix\url{http://dx.doi.org/10.1109/TIT.1982.1056489}

\bibitem{lowe04}
Lowe, D.G.: Distinctive image features from scale-invariant keypoints.
\newblock International Journal of Computer Vision \textbf{60}(2), 91--110
  (2004).
\newblock \doi{10.1023/B:VISI.0000029664.99615.94}.
\newblock \urlprefix\url{http://dx.doi.org/10.1023/B:VISI.0000029664.99615.94}

\bibitem{mclachlan2000}
McLachlan, G., Peel, D.: Finite Mixture Models.
\newblock Wiley series in probability and statistics. Wiley (2000)

\bibitem{Miksik2012}
Miksik, O., Mikolajczyk, K.: Evaluation of local detectors and descriptors for
  fast feature matching.
\newblock In: Pattern Recognition (ICPR), 2012 21st International Conference
  on, pp. 2681--2684 (2012)

\bibitem{perdoch09}
Perd'och, M., Chum, O., Matas, J.: Efficient representation of local geometry
  for large scale object retrieval.
\newblock In: Computer Vision and Pattern Recognition, 2009. CVPR 2009. IEEE
  Conference on, pp. 9--16 (2009)

\bibitem{perronnin07}
Perronnin, F., Dance, C.: Fisher kernels on visual vocabularies for image
  categorization.
\newblock In: Computer Vision and Pattern Recognition, 2007. CVPR '07. IEEE
  Conference on, pp. 1--8 (2007).
\newblock \doi{10.1109/CVPR.2007.383266}

\bibitem{perronnin2015}
Perronnin, F., Larlus, D.: {Fisher Vectors Meet Neural Networks: A Hybrid
  Classification Architecture}.
\newblock In: Proceedings of the IEEE Conference on Computer Vision and Pattern
  Recognition, pp. 3743--3752 (2015)

\bibitem{perronnin10}
Perronnin, F., Liu, Y., S{\`a}nchez, J., Poirier, H.: Large-scale image
  retrieval with compressed fisher vectors.
\newblock In: Computer Vision and Pattern Recognition (CVPR), 2010 IEEE
  Conference on, pp. 3384--3391 (2010).
\newblock \doi{10.1109/CVPR.2010.5540009}

\bibitem{perronin10:improvingFK}
Perronnin, F., S{\`a}nchez, J., Mensink, T.: Improving the fisher kernel for
  large-scale image classification.
\newblock In: Computer Vision - ECCV 2010, \emph{Lecture Notes in Computer
  Science}, vol. 6314, pp. 143--156. Springer Berlin Heidelberg (2010).
\newblock \doi{10.1007/978-3-642-15561-1_11}.
\newblock \urlprefix\url{http://dx.doi.org/10.1007/978-3-642-15561-1_11}

\bibitem{philbin07}
Philbin, J., Chum, O., Isard, M., Sivic, J., Zisserman, A.: Object retrieval
  with large vocabularies and fast spatial matching.
\newblock In: Computer Vision and Pattern Recognition (CVPR), 2007 IEEE
  Conference on, pp. 1--8 (2007).
\newblock \doi{10.1109/CVPR.2007.383172}

\bibitem{philbin08}
Philbin, J., Chum, O., Isard, M., Sivic, J., Zisserman, A.: Lost in
  quantization: Improving particular object retrieval in large scale image
  databases.
\newblock In: Computer Vision and Pattern Recognition, 2008. CVPR 2008. IEEE
  Conference on, pp. 1--8 (2008).
\newblock \doi{10.1109/CVPR.2008.4587635}

\bibitem{razavian2014cnn}
Razavian, A.S., Azizpour, H., Sullivan, J., Carlsson, S.: {CNN} features
  off-the-shelf: an astounding baseline for recognition.
\newblock In: Computer Vision and Pattern Recognition Workshops (CVPRW), 2014
  IEEE Conference on, pp. 512--519. IEEE (2014).
\newblock \doi{10.1109/CVPRW.2014.131}

\bibitem{rublee11}
Rublee, E., Rabaud, V., Konolige, K., Bradski, G.: Orb: An efficient
  alternative to sift or surf.
\newblock In: Computer Vision (ICCV), 2011 IEEE International Conference on,
  pp. 2564--2571 (2011)

\bibitem{salton86}
Salton, G., McGill, M.J.: Introduction to Modern Information Retrieval.
\newblock McGraw-Hill, Inc., New York, NY, USA (1986)

\bibitem{sanchez13}
S{\`a}nchez, J., Perronnin, F., Mensink, T., Verbeek, J.: Image classification
  with the fisher vector: Theory and practice.
\newblock International Journal of Computer Vision \textbf{105}(3), 222--245
  (2013).
\newblock \doi{10.1007/s11263-013-0636-x}.
\newblock \urlprefix\url{http://dx.doi.org/10.1007/s11263-013-0636-x}

\bibitem{sanchez15}
S{\`a}nchez, J., Redolfi, J.: Exponential family fisher vector for image
  classification.
\newblock Pattern Recognition Letters \textbf{59}, 26 -- 32 (2015).
\newblock \doi{http://dx.doi.org/10.1016/j.patrec.2015.03.010}

\bibitem{Simonyan2013}
Simonyan, K., Vedaldi, A., Zisserman, A.: Deep fisher networks for large-scale
  image classification.
\newblock In: C.J.C. Burges, L.~Bottou, M.~Welling, Z.~Ghahramani, K.Q.
  Weinberger (eds.) Advances in Neural Information Processing Systems 26, pp.
  163--171. Curran Associates, Inc. (2013)

\bibitem{simonyan2014}
Simonyan, K., Zisserman, A.: Very deep convolutional networks for large-scale
  image recognition.
\newblock CoRR \textbf{abs/1409.1556} (2014).
\newblock \urlprefix\url{http://arxiv.org/abs/1409.1556}

\bibitem{sivic03}
Sivic, J., Zisserman, A.: Video google: A text retrieval approach to object
  matching in videos.
\newblock In: Proceedings of the Ninth IEEE International Conference on
  Computer Vision, \emph{ICCV '03}, vol.~2, pp. 1470--1477. IEEE Computer
  Society (2003).
\newblock \doi{10.1109/ICCV.2003.1238663}

\bibitem{Smeulders:2000:CIR:357871.357873}
Smeulders, A.W.M., Worring, M., Santini, S., Gupta, A., Jain, R.: Content-based
  image retrieval at the end of the early years.
\newblock IEEE Trans. Pattern Anal. Mach. Intell. \textbf{22}(12), 1349--1380
  (2000)

\bibitem{Sydorov2014}
Sydorov, V., Sakurada, M., Lampert, C.H.: Deep fisher kernels - end to end
  learning of the fisher kernel gmm parameters.
\newblock In: The IEEE Conference on Computer Vision and Pattern Recognition
  (CVPR) (2014)

\bibitem{tolias11:SpeededUp}
Tolias, G., Avrithis, Y.: Speeded-up, relaxed spatial matching.
\newblock In: Computer Vision (ICCV), 2011 IEEE International Conference on,
  pp. 1653--1660 (2011).
\newblock \doi{10.1109/ICCV.2011.6126427}

\bibitem{tolias14}
Tolias, G., Furon, T., J\'egou, H.: Orientation covariant aggregation of local
  descriptors with embeddings.
\newblock In: D.~Fleet, T.~Pajdla, B.~Schiele, T.~Tuytelaars (eds.) Computer
  Vision - ECCV 2014, \emph{Lecture Notes in Computer Science}, vol. 8694, pp.
  382--397. Springer International Publishing (2014)

\bibitem{tolias13:queryExp}
Tolias, G., J{\'e}gou, H.: {Local visual query expansion: Exploiting an image
  collection to refine local descriptors}.
\newblock Research Report RR-8325 (2013).
\newblock \urlprefix\url{https://hal.inria.fr/hal-00840721}

\bibitem{uchida13}
Uchida, Y., Sakazawa, S.: Image retrieval with fisher vectors of binary
  features.
\newblock In: Pattern Recognition (ACPR), 2013 2nd IAPR Asian Conference on,
  pp. 23--28 (2013)

\bibitem{Ullman96}
Ullman, S.: High-Level Vision - Object Recognition and Visual Cognition.
\newblock MIT Press (1996)

\bibitem{Uricchio_2015_ICCV_Workshops}
Uricchio, T., Bertini, M., Seidenari, L., Del~Bimbo, A.: Fisher encoded
  convolutional bag-of-windows for efficient image retrieval and social image
  tagging.
\newblock In: The IEEE International Conference on Computer Vision (ICCV)
  Workshops (2015)

\bibitem{van14}
Van~Opdenbosch, D., Schroth, G., Huitl, R., Hilsenbeck, S., Garcea, A.,
  Steinbach, E.: Camera-based indoor positioning using scalable streaming of
  compressed binary image signatures.
\newblock In: IEEE International Conference on Image Processing (2014)

\bibitem{wang10}
Wang, J., Yang, J., Yu, K., Lv, F., Huang, T., Gong, Y.: Locality-constrained
  linear coding for image classification.
\newblock In: Computer Vision and Pattern Recognition (CVPR), 2010 IEEE
  Conference on, pp. 3360--3367 (2010)

\bibitem{witten99}
Witten, I.H., Moffat, A., Bell, T.C.: Managing gigabytes: compressing and
  indexing documents and images.
\newblock Morgan Kaufmann (1999)

\bibitem{yang09}
Yang, J., Yu, K., Gong, Y., Huang, T.: Linear spatial pyramid matching using
  sparse coding for image classification.
\newblock In: Computer Vision and Pattern Recognition, 2009. CVPR 2009. IEEE
  Conference on, pp. 1794--1801 (2009)

\bibitem{Ng_2015_CVPR_Workshops}
Yue-Hei~Ng, J., Yang, F., Davis, L.S.: Exploiting local features from deep
  networks for image retrieval.
\newblock In: The IEEE Conference on Computer Vision and Pattern Recognition
  (CVPR) Workshops (2015)

\bibitem{ZADB06Similarity}
Zezula, P., Amato, G., Dohnal, V., Batko, M.: Similarity Search: The Metric
  Space Approach, \emph{Advances in Database Systems}, vol.~32.
\newblock Springer (2006)

\bibitem{zhang13}
Zhang, Y., Zhu, C., Bres, S., Chen, L.: Encoding local binary descriptors by
  bag-of-features with hamming distance for visual object categorization.
\newblock In: P.~Serdyukov, P.~Braslavski, S.~Kuznetsov, J.~Kamps, S.~Rüger,
  E.~Agichtein, I.~Segalovich, E.~Yilmaz (eds.) Advances in Information
  Retrieval, \emph{Lecture Notes in Computer Science}, vol. 7814, pp. 630--641.
  Springer Berlin Heidelberg (2013)

\bibitem{zhao13}
Zhao, W., J{\'e}gou, H., Gravier, G.: {Oriented pooling for dense and non-dense
  rotation-invariant features}.
\newblock In: {BMVC - 24th British Machine Vision Conference} (2013)

\bibitem{zhou2014}
Zhou, B., Lapedriza, A., Xiao, J., Torralba, A., Oliva, A.: Learning deep
  features for scene recognition using places database.
\newblock In: Z.~Ghahramani, M.~Welling, C.~Cortes, N.~Lawrence, K.~Weinberger
  (eds.) Advances in Neural Information Processing Systems 27, pp. 487--495.
  Curran Associates, Inc. (2014)

\end{thebibliography}

\appendix
\section{Score vector computation}\label{score vector}
%\section{Score vector computation}
%\label{score vector}

In the following, we have reported the computation of the score function $G_{\lm}^X$, defined as the gradient of the log-likelihood of a data $X$ with respect to the
parameters $\lm$ of a Bernoulli Mixture Model.
Throughout this appendix we have used $\1{\cdot}$ notation to represent the Iverson bracket
which equals one if the arguments is true, and zero otherwise.

Under the independence assumption, the Fisher score  with respect to the generic parameter $\lm_k$ is expressed as:
$
G_{\lm_k}^X =\sum_{t=1}^T \dfrac{\partial\log p(x_t|\lm)}{\partial \lm_k}=
\sum_{t=1}^T \dfrac{1} {p(x_t|\lambda)}\dfrac{\partial}{\partial \lm_k}\left[\sum_{i=1}^K w_i p_i(x_t)\right].
$
To compute $\dfrac{\partial}{\partial \lm_k}\left[\sum_{i=1}^K w_i p_i(x_t)\right]$, we first observe that
\begin{small}\begin{equation}\label{eq:der alpha}
    \begin{split}
    \dfrac{\partial w_i}{\partial \alpha_k}&=
    \dfrac{\partial}{\partial \alpha_k}\left[\dfrac{\exp(\alpha_i)}{\sum_{j=1}^K\exp(\alpha_j)}\right]\\
    &=\dfrac{\exp(\alpha_k)\left(\sum_{j=1}^K\exp(\alpha_j)\right) \1{i=k}-\exp(\alpha_i)\exp(\alpha_k)
    }{\left(\sum_{j=1}^K\exp(\alpha_j)\right)^2}\\
    &= w_k\1{i=k}- w_kw_i
    \end{split}
    \end{equation}\end{small}
and
\begin{small}\begin{equation}\label{eq:der mu}
    \begin{split}
    &\dfrac{\partial p_i( x_t)} {\partial \mu_{kd}}=\dfrac{\partial}{\partial \mu_{kd}}\left[\prod_{l=1}^D \mu_{kl}^{x_{tl}}
    \left(1-\mu_{kl}\right)^{1-x_{tl}} \right]\1{i=k}\\
    &=\left( \1{x_{td}=1}- \1{x_{td}=0}\right)
    \left(\prod_{\substack{l=1 \\ l\neq d}}^D \mu_{kl}^{x_{tl}}\left(1-\mu_{kl}\right)^{1-x_{tl}}\right)\1{i=k} \\
    &=\left( \1{x_{td}=1}- \1{x_{td}=0}\right)\left(
    \dfrac{p_k(x_t)}{\mu_{kd}^{x_{td}}\left(1-\mu_{kd}\right)^{1-x_{td}}}\right)\1{i=k}\\
    &=p_k(x_t)\left( \dfrac{(1-\mu_{kd})\1{x_{td}=1}-\mu_{kd}\1{x_{td}=0}}{\mu_{kd}(1-\mu_{kd})}\right) \1{i=k}\\
    &=p_k(x_t)\left( \dfrac{x_{td}-\mu_{kd}}{\mu_{kd}(1-\mu_{kd})}\right) \1{i=k}.
    \end{split}
    \end{equation}\end{small}
Hence, the Fisher score with respect to the parameter $\alpha_k$ is obtained as
\begin{small}\begin{equation}\label{eq:fs alpha}
    \begin{split}
    G_{\alpha_k}^X &=\sum_{t=1}^T\sum_{i=1}^K\dfrac{ p_i(x_t)} {p(x_t|\lambda)}\dfrac{\partial w_i}{\partial \alpha_k}\stackrel{\eqref{eq:der alpha}}{=}\sum_{t=1}^T \sum_{i=1}^K \dfrac{ p_i(x_t)} {p(x_t|\lambda)}w_k\left( \1{i=k}-w_i\right)\\
    &=\sum_{t=1}^T \left(\dfrac{ p_k(x_t)} {p(x_t|\lambda)}w_k-\sum_{i=1}^K \dfrac{ p_i(x_t)}
    {p(x_t|\lambda)}w_kw_i\right)=\sum_{t=1}^T \left(\gamma_t(k)-w_k\sum_{i=1}^K \gamma_t(i)\right)\\
    &=\sum_{t=1}^T \left(\gamma_t(k)-w_k \right)
    \end{split}
    \end{equation}\end{small}
and the Fisher score related to the parameter $\mu_{kd}$ is
\begin{small}\begin{equation}\label{eq:fs mu}
    \begin{split}
    G_{\mu_{kd}}^X &=\sum_{t=1}^T \dfrac{\partial\log p(x_t|\lm)}{\partial \mu_{kd}}=\sum_{t=1}^T \dfrac{1} {p(x_t|\lm)}\dfrac{\partial}{\partial \mu_{kd}}\left[\sum_{i=1}^K w_i p_i(x_t)\right] \\
    &=\sum_{t=1}^T \dfrac{w_k} {p(x_t|\lm)}\dfrac{\partial p_k(x_t)}{\partial \mu_{kd}}  \stackrel{\eqref{eq:der mu}}{=}\sum_{t=1}^T \dfrac{w_k p_k(x_t)}{p(x_t|\lm)}\left( \dfrac{x_{td}-\mu_{kd}}{\mu_{kd}(1-\mu_{kd})}\right) \\
    &=\sum_{t=1}^T \gamma_t(k)\left(\dfrac{x_{td}-\mu_{kd}}{\mu_{kd}(1-\mu_{kd})}\right).
    \end{split}
    \end{equation}\end{small}

\section{Approximation of the Fisher Information Matrix}\label{FIM}
%\section{Approximation of the Fisher Information Matrix}\label{FIM}
Our derivation of the FIM is based on the assumption (see also \cite {perronnin10,sanchez13}) that for each observation $x=(x_1,\dots,x_D)\in\{0,1\}^D$ the distribution of the occupancy probability
$\gamma(\cdot)=p(\cdot| x,\lm)$ is sharply peaking, i.e. there is one
Bernoulli index $k$ such that $\gamma_x(k)\approx 1$ and $\forall \, i\neq k$, $\gamma_x(i)\approx 0$.
%\comment{lucia}{fare qualche empirical observation?)}.
This assumption implies that
\begin{align*}
\gamma_x(k)\gamma_x(i)\approx  0  &\quad \forall\,k,i=1\dots, K, i\neq k\\
\gamma_x(k)^2\approx  \gamma_x(k) &\quad \forall\, k=1,\dots,K
\end{align*}
and then
\begin{equation}\label{eq: sharp}
\gamma_x(k)\gamma_x(i)\approx\gamma_x(k) \1{i=k},
\end{equation}
where $\1{\cdot} $ is the Iverson bracket.\\
The elements of the FIM are defined as:
\begin{equation}\label{eq:fim}
[F_\lm]_{i,j}=\mathbb{E}_{x\sim p(\cdot|\lm)}\left[\left(\dfrac{\partial\log p(x|\lm)}{\partial \lm_i}\right) \left(\dfrac{\partial\log p(x|\lm)}{\partial \lm_j}\right)\right].
\end{equation}
Hence, the FIM $F_\lm$ is symmetric and can be written as block matrix
\begin{equation*}
F_\lm=\begin{bmatrix}
F_{\alpha,\alpha} & F_{\mu,\alpha}\\
F_{\mu,\alpha}\T  & F_{\mu,\mu}
\end{bmatrix}.
\end{equation*}
By using the definition of the occupancy probability (i.e. $\gamma_x(k)={w_kp_k(x)}/{p(x|\lm)}$) and the fact that $p_k$ is the distribution of a $D$-dimensional Bernoulli of mean $\mu_k$, we have the following useful equalities:  %and straightforward
\begin{small}\begin{align}
    %%\label{eq:u1}&\gamma_x(k)p(x|\lm)=w_kp_k(x) \\
    %\label{eq:u2}&\sum_{x\in\{0,1\}^D}p_k(x)=1 \\
    %\label{eq:u3}&\sum_{x\in\{0,1\}^D}p_k(x)x_{d}=\mu_{kd} \\
    %\label{eq:u4} &\sum_{x\in\{0,1\}^D}p_k(x)x_{d}x_{l}=\mu_{kd}\mu_{kl}\1{d\neq l} +\mu_{kd}\1{d= l}\\
    \label{eq:u5}&\mathbb{E}_{x\sim p(\cdot|\lm)}\left[\gamma_x(k)\right]= \sum_{x\in\{0,1\}^D}\gamma_x(k)p(x|\lm){=}w_k \\
    \label{eq:u6} &\mathbb{E}_{x\sim p(\cdot|\lm)}\left[\gamma_x(k)x_{d}\right]{=}w_k\mu_{kd}\\
    \label{eq:u7}&\mathbb{E}_{x\sim p(\cdot|\lm)}\left[\gamma_x(k)x_{d}x_{l}\right]{=}w_k\mu_{kd}\left(\mu_{kl}\1{d\neq l} +\1{d= l}\right) \\
    \label{eq:u8}&\mathbb{E}_{x\sim p(\cdot|\lm)}\left[\dfrac{\partial\log p(x|\lm)}{\partial \alpha_{k}}\right]\stackrel{ \eqref {eq:fs alpha}}{=}\mathbb{E}_{x\sim p(\cdot|\lm)}\left[\gamma_x(k)-w_{k}\right]%\stackrel{\eqref{eq:u5}}
    {=}0\\
    \label{eq:u9}&\mathbb{E}_{x\sim p(\cdot|\lm)}\left[\dfrac{\partial\log p(x|\lm)}{\partial \mu_{id}}\right]\stackrel{ \eqref {eq:fs mu}}{=}
    \mathbb{E}_{x\sim p(\cdot|\lm)}\left[\dfrac{\gamma_x(k)(x_{d}-\mu_{kd})}{\mu_{kd}(1-\mu_{kd})}\right]%\stackrel{\eqref{eq:u5}+\eqref{eq:u6}}
    {=}0.%
    \end{align}\end{small}
It follows that  $F_\lm$ may approximated by a diagonal block matrix, because the mixing blocks $F_{\mu_{kd},\alpha_{i}}$ are close to the zero matrix:
\begin{small}\begin{equation*}
    \begin{split}
    F_{\mu_{kd},\alpha_{i}}&\quad\stackrel{ }{=}%\eqref{eq:fim}
    \mathbb{E}_{x\sim p(\cdot|\lm)}\left[\left(\dfrac{\partial\log p(x|\lm)}{\partial \mu_{kd} }\right)\left(\dfrac{\partial\log p(x|\lm)}{\partial \alpha_i}\right)\right]\\
    &\stackrel{ \eqref{eq:fs alpha}-\eqref{eq:fs mu}}{=}  \mathbb{E}_{x\sim p(\cdot|\lm)}\left[\gamma_x(k)\dfrac{(x_{d}-\mu_{kd})}{\mu_{kd}(1-\mu_{kd})}(\gamma_x(i)-w_i) \right]\\
    & \quad\stackrel{ \eqref{eq: sharp} }{\approx}\mathbb{E}_{x\sim p(\cdot|\lm)}\left[\dfrac{\gamma_x(k)(x_{d}-\mu_{kd}) }{\mu_{kd}(1-\mu_{kd})}\right]\left(\1{i=k}-w_i\right)\\
    &\quad\stackrel{\eqref{eq:u9}}{=}0.
    \end{split}
    \end{equation*}
\end{small}
The block $F_{\mu,\mu}$ can be written as $KD\times KD$ diagonal matrix, in fact:
\begin{small}\begin{equation}\label{eq:fim mu}
    \begin{split}
    F_{\mu_{id},\mu_{kl}}&\stackrel{ \eqref{eq:fim}}{=}
    \mathbb{E}\left[\left(\dfrac{\partial\log p(x|\lm)}{\partial \mu_{id} }\right)\left(\dfrac{\partial\log p(x|\lm)}{\partial \mu_{kl}}\right)\right]\\
    &\quad\stackrel{ \eqref {eq:fs mu}}{=}\mathbb{E}_{x\sim p(\cdot|\lm)}\left[\gamma_x(i)\gamma_x(k)\dfrac{(x_{d}-\mu_{id})}{\mu_{id}(1-\mu_{id})}\dfrac{(x_{l}-\mu_{kl})}{\mu_{kl}(1-\mu_{kl})} \right]\\
    %&\quad\stackrel{ \eqref{eq: sharp 1} }{\approx} \dfrac{\mathbb{E}_{x\sim p(\cdot|\lm)}\left[\gamma_x(k)^2(x_{d}-\mu_{kd})(x_{l}-\mu_{kl})\right]}{\mu_{kd}\mu_{kl}(1-\mu_{kd})(1-\mu_{kl})}\1{i=k}\\
    &\quad\stackrel{ \eqref{eq: sharp} }{\approx}\mathbb{E}_{x\sim p(\cdot|\lm)}\left[ \dfrac{\gamma_x(k)(x_{d}-\mu_{kd})(x_{l}-\mu_{kl})}{\mu_{kd}\mu_{kl}(1-\mu_{kd})(1-\mu_{kl})}\right]\1{i=k}\\
    %&\quad= \mathbb{E}_{x\sim p(\cdot|\lm)}\left[\dfrac{\gamma_x(k)(x_{d}x_{l}-\mu_{kl}x_{d}-\mu_{kd}x_{l}+\mu_{kd}\mu_{kl})}{\mu_{kd}\mu_{kl}(1-\mu_{kd})(1-\mu_{kl})}\right]\1{i=k}\\
    &\stackrel{\eqref{eq:u5}-\eqref{eq:u7}}{=} \dfrac{ w_k(\mu_{kd}\mu_{kl}\1{d\neq l} +\mu_{kl}\1{d= l}-\mu_{kd}\mu_{kl})}{\mu_{kd}\mu_{kl}(1-\mu_{kd})(1-\mu_{kl})}\1{i=k}\\
    &\quad= \dfrac{ w_k(\mu_{kd}\1{d\neq l} +\1{d= l}-\mu_{kd})}{\mu_{kd}(1-\mu_{kd})(1-\mu_{kl})}\1{i=k}\\
    &\quad= \dfrac{  w_k}{\mu_{kd}(1-\mu_{kd})}\1{i=k}\1{d=l}.
    \end{split}
    \end{equation}\end{small}
The relation \eqref{eq:fim mu} points that the diagonal elements  of our FIM approximation %$F_{\mu,\mu}$ 
are
$
{  w_k}/ \mu_{kd}(1-\mu_{kd}) % F_{\mu_{kd},\mu_{kd}}=
$
and the corresponding entries in
$L_\lm$ (i.e. the square root of the inverse of FIM) equal $ \sqrt{{\mu_{kd}(1-\mu_{kd})}/{ w_k}}$.
The block related to the $\alpha$ parameters is
$
F_{\alpha,\alpha}= (\text{diag}(w)-ww\T )
$
where $w=[w_1,\dots,w_K]\T $, in fact:
\begin{small}\begin{equation*}
    \begin{split}
    F_{\alpha_{k},\alpha_{i}}&\quad\stackrel{ \eqref{eq:fim}}{=}
    \mathbb{E}_{x\sim p(\cdot|\lm)}\left[\left(\dfrac{\partial\log p(x|\lm)}{\partial \alpha_{k} }\right)\left(\dfrac{\partial\log p(x|\lm)}{\partial \alpha_i}\right)\right]\\
    &\quad\stackrel{ \eqref{eq:fs alpha}}{=}\mathbb{E}_{x\sim p(\cdot|\lm)}\left[(\gamma_x(k)-w_k)(\gamma_x(i)-w_i) \right]\\
    %&\quad=\mathbb{E}_{p(\cdot|\lm)}\left[\gamma_x(k)\gamma_x(i)- \gamma_x(k)w_i-\gamma_x(i)w_k +w_iw_k \right]\\
    &\quad\stackrel{ \eqref{eq: sharp} }{\approx}\mathbb{E}_{p(\cdot|\lm)}\left[\gamma_x(k)\1{i=k}- \gamma_x(k)w_i-\gamma_x(i)w_k +w_iw_k \right]\\
    &\stackrel{\eqref{eq:u5}-\eqref{eq:u6}}{=}\left( w_k\1{i=k}-w_iw_k \right).
    \end{split}
    \end{equation*}\end{small}
The matrix  $F_{\alpha,\alpha}$ is not invertible (indeed $F_{\alpha,\alpha}\mathbf{e}={0}$ where $\mathbf{e}=[1,\dots,1]\T $)
due to the dependence of the mixing weights ($\sum_{i=1}^K\alpha_i=\sum_{i=1}^K w_i=1$).
Since there are only $K-1$ degrees of freedom in the mixing weight, as proposed in \cite{sanchez13}, we can fix $\alpha_K$
equal to a constant without loss of generality and work with a reduced set of $K-1$ parameters: $\tilde{\alpha}=[\alpha_1,\dots,\alpha_{K-1}]\T $.

Taking into account the Fisher score with respect to $\tilde{\alpha}$,
i.e.
\begin{equation*}
G_{\tilde{\alpha}}^X= \nabla_{\tilde{\alpha}}\log p(X|\lm)=[G_{{\alpha_1}}^X,\dots, G_{{\alpha_{K-1}}}^X]\T =\widetilde{G_{\alpha}^X},
\end{equation*}
the corresponding block of the FIM is
$
F_{\tilde{\alpha},\tilde{\alpha}}= (\text{diag}(\tilde{w})-\tilde{w}\tilde{w}\T ),
$
where $\tilde{w}=[w_1,\dots,w_{K-1}]\T $.
The matrix $F_{\tilde{\alpha},\tilde{\alpha}}$ is invertible, indeed it can be decomposed
into a product of an invertible diagonal matrix $D=\text{diag}(\tilde{w})$ and an invertible elementary matrix
\footnote{ An elementary matrix $E(u,v,\sigma)=I-\sigma uv^H$ is non-singular if and only if $\sigma v^Hu\neq1$ and in this case the inverse is
    $E(u,v,\sigma)^{-1}=E(u,v,\tau)$ where $\tau=\sigma/(\sigma v^Hu-1)$. More details on this topic can be found in \cite{householder64}.  }
$E(\mathbf{e},\tilde{w},-1)= I-\mathbf{e}\tilde{w}\T $; its inverse is
\begin{equation*}
\begin{split}
F_{\tilde{\alpha},\tilde{\alpha}}^{-1}&=\text{diag}(\tilde{w})^{-1}\left(I+\dfrac{1}{\sum_{i=1}^{K-1}w_i-1}\mathbf{e}\tilde{w}\T \right)=
\left(\text{diag}(\tilde{w})^{-1}+\dfrac{1}{w_K}\mathbf{e}\mathbf{e}\T \right).
\end{split}
\end{equation*}
It follows that
\begin{equation*}
\begin{split}
K_{\tilde{\alpha}}(X,Y)&=(G_{\tilde{\alpha}}^X)\T  F_{\tilde{\alpha},\tilde{\alpha}}^{-1} G_{\tilde{\alpha}}^Y= \left((G_{\tilde{\alpha}}^X)\T \text{diag}(\tilde{w})^{-1}G_{\tilde{\alpha}}^Y+\dfrac{1}{w_K}(\mathbf{e}\T G_{\tilde{\alpha}}^X)(\mathbf{e}\T G_{\tilde{\alpha}}^Y) \right)=\sum_{k=1}^K \dfrac{G_{{\alpha_k}}^X G_{{\alpha_k}}^Y}{ w_k}
\end{split}
\end{equation*}
where we used
$
\mathbf{e}\T  G_{\tilde{\alpha}}^Z=\sum_{k=1}^{K-1}\sum_{z\in Z} \left(\gamma_{z}(k)-w_k \right)
=-\sum_{z\in Z} \left(\gamma_{z}(K)-w_K\right)=-G_{{\alpha_K}}^Z.
$\\
By  defining
$
\mathcal{G}_{\alpha_k}^X =\dfrac{1}{\sqrt{w_k}}\sum_{x\in X} \left(\gamma_x(k)-w_k\right),
$
we finally obtain
$
K_{\tilde{\alpha}}(X,Y)=\left(\mathcal G_{\alpha}^X\right)\T \mathcal G_{\alpha}^Y.
$
Please note that we don't need to explicitly compute the Cholesky decomposition of the matrix $F_{\tilde{\alpha},\tilde{\alpha}}^{-1}$  because
the Fisher Kernel $K_{\tilde{\alpha}}(X,Y)$ can be easily rewritten as dot product between the feature vector $\mathcal G_{{\alpha}}^X$ and $\mathcal G_{{\alpha}}^Y$.
\end{document}